# Mapping Big Data into Knowledge Space with Cognitive Cyber-Infrastructure

Hai Zhuge

*Chinese Academy of Sciences, Beijing, China*
*Aston University, Birmingham, B4 7ET, UK*

**Abstract**

*Big data research has attracted great attention in science, technology, industry and society. It is developing with the evolving scientific paradigm, the fourth industrial revolution, and the transformational innovation of technologies. However, its nature and fundamental challenge have not been recognized, and its own methodology has not been formed. This paper explores and answers the following questions: What is big data? What are the basic methods for representing, managing and analyzing big data? What is the relationship between big data and knowledge? Can we find a mapping from big data into knowledge space? What kind of infrastructure is required to support not only big data management and analysis but also knowledge discovery, sharing and management? What is the relationship between big data and science paradigm? What is the nature and fundamental challenge of big data computing? A multi-dimensional perspective is presented toward a methodology of big data computing.*

## 1. Introduction

Data represent things. Sensing, storing, processing, managing, analyzing and explaining data have become an important way to know the statuses and behaviors of an observed system. Various data analytic tools are being developed to answer queries, discover patterns in data, help test hypotheses, and make predictions and decisions.

*1.1 Big data surge*

Big data has become a hot topic not only in science and engineering but also in business and social sectors. With intensive discussion in scientific journals and various social medias, public has been impressed that big data with statistical tools offer a powerful approach to solving hard problems. Some people even suggest that science can advance without models and theories (Anderson, 2008).

Almost all domains such as social management (e.g., smart cities), business (e.g., manufacturing and marketing), education, health, security, environment (e.g., monitoring, protection, and industrial symbiosis, etc.), energy, and sciences (especially in physics, astronomy, biology, ecology, and earth science) have raised the challenge of accessing, managing and analyzing rapidly expanding big data, especially the dispersed and heterogeneous nature of these data (Howe, 2008) (Reichman, et al. 2011). On one hand, it is getting harder and harder to manage and make use of big data without the help of new tools and methods. On the other hand, it is getting harder and harder to modeling complex systems that generate big data.

Business sectors have paid more and more attention to the timely analysis of big data for strategic and operational decisions. However, big data analysis can only be effective when data analyzers have the knowledge of business strategies, and can link the potential insights to business opportunities. Therefore, chief information officers need to share knowledge with the heads of business departments to find the business opportunities indicated by big data. In addition to value, organizational capabilities are important for business. It is necessary to move analytics from data into core business (Davenport, et al. 2012).

Big data have already enabled Google search, translation and social network. Google processes over 24 petabytes every day. However, successful big data applications like Translation (https://translate.google.com) and Flu Trends (http://www.google.org/flutrends/) indicate that big data provide an implication of facts (e.g., real meaning and real flu) rather than knowledge. Some researchers have different views on big data (Marcus and Davisapril, 2014).

*1.2 Understanding and initiatives*

There are different understandings on big data. Public understanding is that the volume of data is too big to be processed by current software and hardware. Linear algorithms are not suitable for extremely big data. The comprehensive understanding on big data includes not only volume but also such features as velocity (streaming data), variety (heterogeneous data), and veracity (noisy, inaccurate and unclean data) (Gartner, et al; McAfee and Brynjolfsson, 2012). These understandings concern data itself.

IBM big data initiative is to integrate current techniques such as database, cloud, stream processing and content computing to upgrade enterprise information systems toward higher performance and better services for enterprise management and decision. For example, data management research and warehouse research aim at obtaining industry-leading database performance while lowering administration, storage, development and server costs, and realizing extreme speed with capabilities optimized for analytics workloads. The stream computing research aims at efficiently delivering real-time analytic processing on constantly changing data and enabling descriptive and predictive analytics to support real-time decisions. The content management research enables comprehensive content lifecycle and document management with cost-effective control of existing and new types of content with scale, security and stability. This enterprise computing strategy significantly upgrades enterprise information systems (http://www.ibm.com/big-data). This is the significant extension of traditional data warehouse (Inmon, 2005), which is to integrate, organize and manage data from multiple sources for supporting estimation and decision.

Some researchers proposed to develop big data research toward a science (data science), aiming at deriving valuable insights (or knowledge) from big data. Insights from big data enable enterprises to make better decisions in deepening customer engagement, optimizing operations, preventing threats and fraud, etc. Researchers from various disciplines such as statistics, business, social science, communication, and computer science are investigating big data from various aspects such as modeling, analysis, mining, management and utilization. Predictive knowledge helps make insightful decision in business applications (Dhar, 2013).

The converging effort from multiple disciplines and diverse communities is becoming an important driven force of pushing forward big data research.

1.3 *Data as computing*

How to efficiently organize and process data is a persistent research issue at every development stage of computing. Data were structured specific to algorithms (Bachman,1969; Wirth, 1978), organized in a way loosely dependent on algorithms in databases (Codd, 1982), encapsulated into objects with interfaces and operations in object-oriented programming and modeling (Jacobson, 1999), and organized in a way independent on algorithms just as Web pages independent on search engines (Berners-Lee, 2000; Brin and Page, 1998), and distributed onto decentralized algorithms run on massive machines scheduled for the efficiency of communication, storage, retrieval, and processing like indexing large-scale Web pages crawled constantly (Dean and Ghemawat, 2008). These changes accompany the paradigm shift of software and hardware development.

With the wide use of the Internet and various devices such as scientific instruments, sensors and mobile phones, data are being constantly collected through specially designed devices to help explore various complex systems.

1.4 *Big networks*

The World Wide Web, the biggest artificial network in human history, contains free texts, webpages, databases, and online social networks, which change from time to time. Research like information retrieval has been working on processing big data on the evolving Web. Searching big data on the Web has gradually changed people's way of memory (Sparrow, et al, 2011). Big data on the Web enable some problem-solving tasks to be transformed into a search problem — people have been used to searching the solutions on the Web for many daily-life problems or technical problems. Experiencing easy search, *human is changing the way of thinking toward interactive thinking*.

Web scientists proposed Web Science as a study of the large-scale socio-technical systems, involved in the relationship between people and technology, the ways that society and technology co-constitute one another and the impact of this co-constitution on broader society (Berners-Lee, et al., 2006).

Many social networks such as Facebook, Twitter and LinkedIn based on the Web are continually generating big data from human. Social networks are evolving communities with massive participation of people. Scientists often involve in collaborative networks for sharing scientific data and research results. Data in social networks change more often than webpages, which make real-time analysis more important in applications.

Social networks form communities in society, which divides people and data they generate and use with the evolution of the communities. This requests data management system to manage not only data but also the networks that generate and use data. This indicates a *locality principle of managing data: dealing with big data by localizing data to the people who generate and use data*.

The arising problem is that social networks run on different infrastructures operated by different companies. Currently, users cannot share data between different social networks. This problem involves in *the global accessibility between data*, which is beyond the volume, velocity, variety and veracity features of big data because it concerns infrastructure.

The growing scale and variety of networks such as online social networks, the Internet of Things, and mobile communication networks raise *the challenge of managing not only big data but also the big networks that generate big data*. Processing different networks need to consider different scales and characteristics of nodes, links and operation rules when using general graph-based methods. Many graph-based methods like community discovery algorithms are not suitable for big networks.

1.5 *The shift of science paradigm*

Turing Award laureate Jim Gray described the fourth paradigm of scientific research as a fundamental shift of scientific research from the model-driven paradigm to the data-driven paradigm (Gray, 2007). Science has experienced the following development process: early experience period → theoretical period (through modeling and generalization) → computational period (simulating complex phenomena) → data exploration (new methodology integrating theory, experiment, and simulation).

Hilbert and Turing are distinguished representatives of the theoretical research paradigm in $20^{th}$ century. In 1921, Hilbert advocated to establish mathematics on a solid and provably consistent foundation of axioms, from which, all mathematical truths could be logically deduced. He formulated Entscheidungsproblem (i.e., decision problem) in 1928: *Could an effective procedure be designed to demonstrate in a finite time whether any given mathematical proposition is, or is not, provable from a given set of axioms*? Gödel, Church and Turing proved that any consistent axiomatic theory sufficiently rich to enable the expression and proof of basic arithmetic propositions could be neither complete (Gödel) nor effectively decidable (Church and Turing). Turing machine used for solving the decision problem became the fundamental computing model and it has influenced computing for over sixty years.

Data-intensive science exploration consists of three basic activities: *data capture through devices or simulator, curation through software, analysis through data* *management and statistics, and visualization*. Scientists can access data in understandable forms to support problem-solving, observation, verification, and activity preservation.  Scientists in more and more areas such as physics, astronomy, and biology are used to explore data generated by various devices rather than directly observing the nature.

The fourth paradigm was further explained by Gray's colleagues and other researchers (Hey et al, 2009; Bell, Hey and Szalay, 2009). The long-term data provenance and community access to distributed scientific data in various medias are basic work for data-intensive exploration.

Collecting scientific data through various devices and discovering scientific principles by analyzing data has been adopted by scientists as empirical research approach.  An example is that Johannes Kepler discovered the laws of planetary motion by analyzing Tycho Brahe's catalog of observational data.

When people are difficult to understand a complex system, simplification is often used to build a model.  High-performance computers enable complex systems to be simulated with more variables.  This leads to the computational paradigm of scientific research. Computational models provide a means for people to observe, analyze and understand the observed system. Models can be tested by data and improved to better reflect the nature.  The fourth paradigm of scientific research is the development of the computational paradigm according to Gray.

In Web age, scientists have used search engines and online digital resources (including scientific data and papers) and online social networks frequently when doing research, especially at the stages of writing proposal (e.g., checking related idea, searching collaborators and references), collecting data, analyzing data, and writing papers (citation, submission, review, etc.).

For data-intensive scientific exploration, inventing new devices plays an important part of discovering principles and rules in the new observed system.  So, the fourth generation research is not only about big data exploration but also about the devices that generate data.  *The invention of devices is basic because the function and the working principles of devices determine the explanation ability of the generated data*. In science history, the invention of new instruments such as Hubble Space Telescope and Magnetic Resonance Imaging often brings great scientific progress.

1.6 *Big data meets the fourth industrial revolution*

Following the mechanization of production using steam power, mass production using electric power, and automatic production based on IT, the fourth industry revolution will realize intelligent manufacturing.

Initiatives include the German's high-tech strategy industry 4.0 program that focuses on adaptability, resource efficiency and human factors in business and value processes (Roland Berger, 2014), the USA's Smart Manufacturing Leadership Coalition that promotes collaborative R&D with shared infrastructure that facilitates the broad adoption of intelligent manufacturing (https://smartmanufacturingcoalition.org), the Industrial Internet projects that bring together two revolutions: multi-disciplinary innovations arising from the industrial revolution, innovations from computing, information and communication systems brought by Internet revolution (http://www.industrialinternetconsortium.org), and Intelligent Manufacturing Systems that focus on frontier technologies in predictive analytics including prognostics and health management technologies, cyber-physical systems, industrial big data analytics, and intelligent decision (http://www.imscenter.net). Big Data will help predict the possibility to increase productivity, quality and flexibility and thus to understand the advantages of competition.

The globalization of industry requests the *accessibility of global data on demand*.

*1.7 Observed system, data, knowledge, human and machine*

The relations between the observed system, knowledge, data, human and machine are depicted in Figure 1. Human collect the data from the observed system through observation with a certain equipment, aim and knowledge. Analyzing data helps human to know the status of the observed system and discover its rules. Data can also help verify rules and generate rules as knowledge. People can also directly analyze the data with aim and knowledge to generate knowledge by proposing and verifying assumptions.

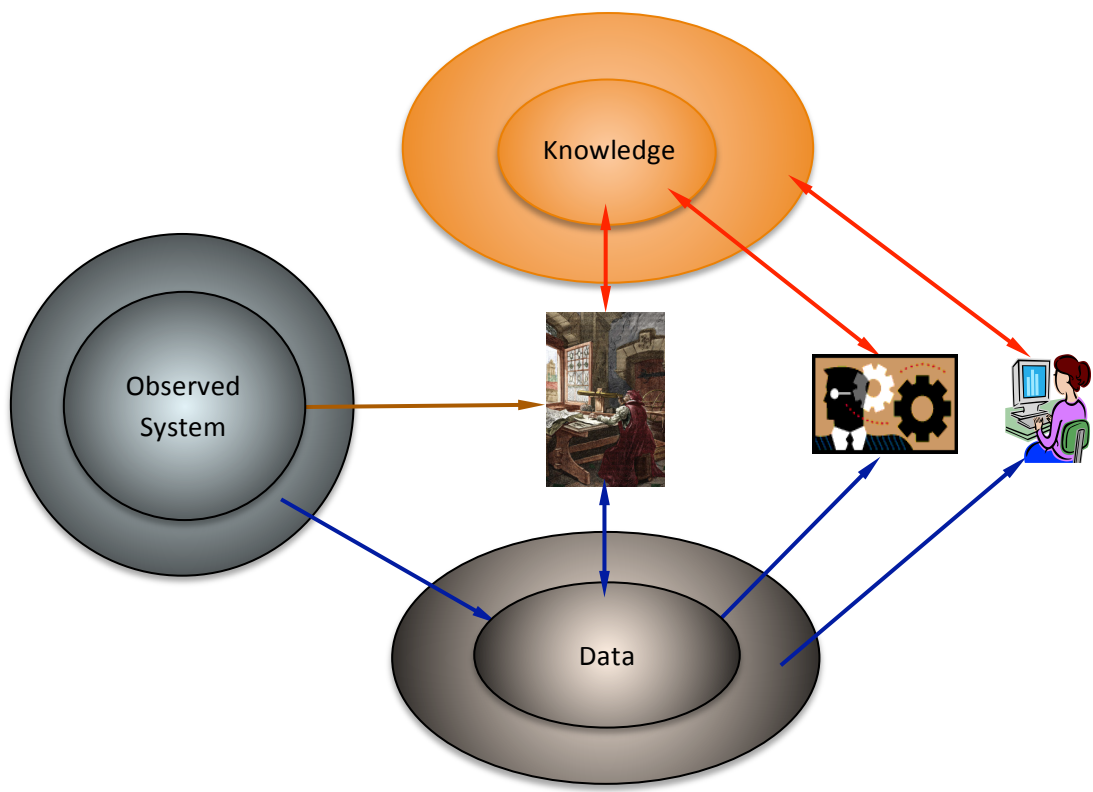

Figure 1. Observed system, data, knowledge, human and machine.

The observed system, data and knowledge evolve during observation, discovery and knowledge processes. The new computing infrastructure helps people to detect, store, manage and analyze big data.  Human can know more status and rules on the observed system with the help of advanced computing infrastructures.

The growing big data provide plentiful resources for people to know, understand and predict the changing world.  Globally shared scientific data can indeed speed up research process and accelerate the development of sciences. However, most current research and development on big data use existing technologies with new conditions.  Big data research needs a new methodology to guide fundamental research and applications.

1.8 *Questions*

Tracing the origin of big data research and linking big data to behavior, knowledge and cyber-infrastructure is a way to explore big data methodology. Answering the following questions helps explore big data from science, system, information and data perspectives.

(1) What is big data? In addition to existing definitions, we hope to find a more fundamental answer, which concerns the following aspects.
   a) What are the basic methods for representing, managing and analyzing big data?  Section 2, section 3 and section 4 discuss this issue.
   b) What are the capability and limitation of big data?  Section 9 discusses this issue.
   c) What is the nature of big data computing?  Section 9 discusses this issue. The conclusion section summarizes the understanding of big data.
(2) What is the relationship between big data and knowledge?  Section 5 discusses this issue.
   a) What is knowledge?  This fundamental problem is necessary to be revisited in big data research because tremendous efforts have been made to extract knowledge from data and many researchers aim at extracting knowledge from big data.  Consequently, we have the following questions:
   b) Can knowledge be discovered in big data?
   c) What is between representation and knowledge?
   d) How to discover and effectively manage knowledge?
(3) What kind of cyber-infrastructure is required to manage big data and to support data analysis, discovery of knowledge, decision and scientific research? Section 6 will discuss this issue with the following two questions.
   a) Can big data be mapped into knowledge space through an intelligent cyber-infrastructure?
   b) How to make use of the state-of-the-art technologies to realize intelligent cyber-infrastructure?
(4) What is the relationship between big data and science paradigm?  Section 7 and section 8 discuss this issue.
(5) What is the fundamental challenge of big data computing? Section 9 discusses this issue.

## 2 Semantic Link Network — Weaving Implicit Web on Various Data

2.1 *Modeling with local views and global views*

Representation of various data is the basis of effectively operating data. Different representations support different operations. One strategy of representation is global definition. For example, relational databases use relational algebra as the global representation to organize data. Another strategy of representation is local-to-global formation. Local representations connect one another to render a global representation. The global definition is generally fix and it requires insight on the global data. The local-to-global strategy does not rely on fix representation but dynamically renders the semantics of data through constant interactions with operations on data.

The World Wide Web uses URL (Uniform Resource Locator) to connect local resources (including Web pages and data) to render a global representation. The URL-based hyperlinks self-organize Web resources to demonstrate some basic characteristics like small world (Albert, 1999), and transmit influence of Web resources through the links.

The linked data is a method of enabling data on the Web from different sources to be connected and queried based on Web standards such as Resource Description Framework (RDF, www.w3.org/RDF/), Web Ontology Language (OWL, www.w3.org/TR/owl-features/) and SPARQL query language for RDF (www.w3.org/TR/rdf-sparql-query/).

Establishing ontology is a global definition strategy. It usually relies on human definition. Open online contents like Wikipedia provide the sources for automatically extracting ontology. Research on semiotics and grounding symbols is to enrich the semantics of symbols (Roy, 2005).

Big data from different systems may need different representations and computing models. Sometimes it is hard to find a unified representation for different representations. Data integration was to solve the problem of combining heterogeneous data from different sources, and the problem of providing users with a unified view on these data (Lenzerini, 2002; Halevy, 2006). IBM regards data integration as the combination of technical and business processes used to combine data from disparate sources into meaningful and valuable information. The key is to select appropriate data sources and establish a mapping from various data structures into a required representation. Network visualization has been used to realize data integration (Smoot, 2011). It is not hard to integrate data with known structures and operations. *The challenge is to integrate big data from changing sources with unknown structures*. The key to solve the problem is the establishment, recognition, derivation and integration of semantics on data.

Big data generated by a complex system reflect the status and behavior of the system. So, it requests a representation that can help human recognize the complex system that generates big data. It is necessary to think what is the fundamental semantics that human understand and use.

Semantics is originated from senses, linking senses, categorization of senses,

and generalization or specialization of categories. Semantic links are clues that need human to discover, and they can be regarded as a specialization of the category of binary relation that widely exists in the real world. Some semantic links are explicit like citation and coauthor in scientific papers while others are implicit like implicit citation in literature. It is non-trivial to discover some implicit links since it may need scientific research. Statistics and machine learning methods can help identify possible indicators (e.g., words) but they are not enough to discover implicit links in data. Heuristic rules are necessary for guiding the discovery of various implicit links.

A semantic link network models the complex system with global definition and local-to-global formation. The global view of a semantic link network consists of a set of semantic nodes (representing anything), a set of semantic links between nodes, and a semantic space. Semantic nodes can be a category or an instance with attributes and ranks. Explicit semantic links are created by explicit citation between nodes. Implicit semantic links are implicated by the attributes of nodes (e.g., blood relation) or interactions (e.g., friendship) between nodes. The semantic space includes a hierarchy of categories and a set of rules for reasoning and inferring semantic links, influencing nodes and links, networking, and evolving the network (Zhuge, 2009). The local-to-global formation mechanism forms implicit patterns through various interactions between objects. The function of the semantic link network can be simulated by object-oriented methods, multi-agents, and service-oriented architecture (Zhuge, 2011).

Social semantics is represented through interaction behaviors in social context (Zhuge, 2010). Interaction can be human-human interaction, human-data interaction, and machine-data interaction, which can derive more interactions such as human-data-human interaction and machine-data-machine interaction. For example, on emerging an important statement when writing a paper, researchers often search the Web to see whether it has been done or not, if it has been done, researchers often hope to find those most relevant and important papers, and researchers often search again to see whether there are other relevant works of the authors or not. A decision of citing relevant papers is made through searching, reading, and selecting behaviors on the same set of papers. A citation decision may be changed during revision. The revision of citation or the removal of citation reflects the author's thinking process. Different researchers may make different decisions on the same set of papers. The semantics of these papers is reflected by not only texts but also behaviors. Previous research on semantics (e.g., semantic web) neglects this issue.

*Semantic links connect various categories through conscious or unconscious interactions, reflecting the rules of the system that generates big data*.

A semantic link network can be huge as it can grow unlimitedly with continually adding nodes and links to the network and its nodes can contain big data. The growing network continually changes the ranks of nodes and links. A semantic link network may evolve communities, following different rules and rendering different behaviors.

A semantic link network is also a tool that can help analyze complex system with the interaction between local views and global views through the evolution. The semantic link network has been extended to link various objects following different rules in cyberspace, physical space and social space (Zhuge, 2011; Zhuge, 2012).

2.2 *The big gap*

Statistics has become hot in recent ten years in computing applications. With wide use of statistics in processing texts, words are commonly used as the basic objects of many models such as word frequency (TF-IDF), vector space model (Salton, et al., 1975), latent semantic analysis (Deerwester, et. al., 1990; Hofmann, 1999), and topic model (Blei, 2012). These models could reflect writers' habit of using words in large document sets, but they neglect the fact that isolated words neither provide semantics for computers to derive nor for human to understand. For example, it is difficult for people to understand the meaning of a topic generated by the topic model, and computers cannot explain a topic and distinguish the meaning between topics, for example, between <tiger, eat, cat> and <cat, eat, tiger>, and between <policeman, arrest, thief> and <thief, arrest, policeman>. *There is a big gap between words, languages and language using*.

Actually, computing involves in two spaces: cyberspace where data is stored, computed and transmitted, and social space where human use computers and interact with each other. There is a big gap between the representations in the two spaces. Clarifying the gap between the spaces, and establishing the mapping between the representations in different spaces is a way to enable both computer and human to maximize their abilities.

Relational databases provide SQL (Structured Query Language) for programmers to understand data structure. However, SQL is limited in ability to represent the meaning as in human writings. Search engines enable users to use their words to retrieve webpages indexed under the words extracted from the crawled webpages. However, current search engines are limited in ability to support exact retrieval and question answering. The key issue is the gap between meaning and words, and the mismatching between the diverse meanings of users (readers), the diverse meanings of writers, and the diverse words they used. Object-oriented programming can be regarded as an effort to unify the representation of computing objects and the representation of real-world objects (Jacobson, 1999), but it mainly concerns generalization and specialization of classes. The generative model is to simulate the process of understanding and writing (Wittrock, 1974). The formation process of a system reflects a certain meaning but it is hard to represent the whole meaning of a system.

2.3 *The levels of representation*

The World Wide Web minimizes its global definition to free links: one HTML-based Webpage or any part of it can cite another webpage using URL, and browsers can display webpages in natural languages for ordinary users to read and navigate with

URL. The HTML is for programmers to use. The IP address bind to URL is for the Internet to use.   However, IP and URL are irrelevant to the meaning of the content of webpage. This limits the ability of browsers and search engines. The Extensible Markup Language (XML) tries to use a tree structure to organize resources so as to enable both human and machine to process the content based on understanding. However, this tree structure is not enough to represent human meaning.   HTML 5 (http://www.w3.org/TR/html5/) integrates class, semantic element and data operation to enable future browsers to know more about the semantics of the resources while displaying rich contents.   It supports links between contents rather than the links on addresses.

Semantic link network uses multiple levels of representations:

(1) Representation for human to read (denoted as *Rep*(*H*) in table 1), which enables human to easily understand meaning, for example, a summary in natural language and image.
(2) Representation for computers to process and calculate (denoted as *Rep*(*C*) in table 1), which enables computers to store and compute, for example, data type and data structures.
(3) The representation of knowledge.

Four tables can be used to represent a semantic link network. Table 1 is the representation of semantic links (denoted as $L_1$, $L_2$, …, etc.).   Table 2 represents semantic nodes (denoted as $N_1$, $N_2$, …, etc.).   Table 3 represents the connections between semantic nodes. Table 4 represents the rules for deriving new semantic links from existing links (denoted as $R_1$, $R_2$, ..., etc.).   The four tables expand with the evolution of the network.   These tables represent a global view but they may not be globally maintained in practice.

Table 1. Semantic links.

| ID | *Rep*(*C*) | Word | *Rep*(*H*) | *Rep* (*K*) |
|---|---|---|---|---|
| $L_1$ | > | Greater | *X* is greater than *Y* in number | Number |
| $L_2$ | = | Equal | *X* is equal to *Y* in number | Number |
| $L_3$ | String: Publish | Publish | *X*'s writing *Y* is printed and publicized | Write, publish |
| $L_4$ | Sting: Cite | Cite | *X*'s published writing uses *Y*'s published writing | reference |
| …… | …… | …… | …… | …… |

Table 2. Semantic nodes.

| ID | *Rep* (*C*) | Word | *Rep* (*H*) | *Rep*(*K*) |
|---|---|---|---|---|
| $N_1$ | String: Turing | Turing | British mathematician, pioneer of computer science Summary of Turing's biography in Wikipedia. | Turing machine Intelligent machine Turing test |
| $N_2$ | String: Bush | Neumann | Pioneer of computer system architecture.   Summary of Neumann's biography in Wikipedia. | Computer system architecture |

| $N_3$ | Sting: Gray | Gray | Pioneer of database Summary of Gray's biography in Wikipedia. | Database, transaction |
| $N_5$ | File: Bush-paper.text | Bush, Think | Paper: V. Bush, As we may think, Atlantic Monthly, July (1945) 101–108. Summary of paper: It foresaw multi-media and the web, and proposed Memex for the first time. …… | Information storage and information retrieval |
| $N_6$ | Class: Computer | Computer | A general-purpose programmable machine. Summary of computer according to Wikipedia. | Hardware, Software |
| …… | …… | …… | …… | …… |

Table 3. Connection between nodes.

|  | $N_1$ | $N_2$ | $N_3$ | $N_4$ | …… |
|---|---|---|---|---|---|
| $N_1$ | Link set | Link set | Link set | Link set | …… |
| $N_2$ | Link set | Link set | Link set | Link set | …… |
| $N_3$ | Link set | Link set | Link set | Link set | …… |
| $N_4$ | Link set | Link set | Link set | Link set | …… |
| …… | …… | …… | …… | …… | …… |

Table 4. Rules for reasoning.

| ID | *Rep* (*C*) | Word | *Rep* (*H*) | *Rep* (*K*) |
|---|---|---|---|---|
| $R_1$ | If ($L_1 \in (N_1, N_2)$ & $L_2 \in (N_2, N_3)$), then add $L_3$ to ($N_1, N_3$), in Table 3 | Cite, co-occur | If *A* cites *C*, and *B* cites *C*, then *A* and *B* are on the same topic. | Cite, Occur, Topic |
| $R_2$ | If $N_1=N_2$ and $N_2=N_3$, then $N_1=N_3$ | Equal | If *A* is identical to *B*, and *B* is identical to *C*, *A* is identical to *C*. | Transitivity |
| …… | …… | …… | …… | …… |

Computers simply process relations (e.g., "publish") and names (e.g., "Turing") as strings or other types. Isolated words are neither suitable for computers to understand nor for humans to understand. Understanding a representation requires users to have knowledge (e.g., the knowledge of number), which is the basis for understanding and representation. For simplification, the representation of knowledge (*Rep*(*K*)) is denoted as a link in the tables. The rest part of this paper will discuss knowledge in detail.

*2.4 Discovering semantic link*

Human intelligence is based on two parts: the mind and the diverse links in the external world. The recognition of the implicit links helps establish insight and make prediction. Knowing the implicit semantic links and relevant rules, computing systems can help human to recognize more implicit links by discovering rules, managing and predicting semantic links, and supporting analysis (Zhuge, 2012).

Different from hyperlink analysis (Kleinberg, 1999; Liben-Nowell and Kleinberg, 2007; Clauset, et al., 2008), semantic link analysis concerns more aspects, including:

(1) *Structure*. How to model various semantic nodes, including user behaviors in a particular application? What are the characteristics of semantic nodes, semantic links and the formation process of a semantic link network?

(2) *Rules*. What are the rules for linking one semantic node to the other, for reasoning on semantic links, for explaining a link, a node and a semantic community, and for estimating the effect of operations and the trend of evolution?

(3) *Principle*. What are the principles of the semantic link network? e.g., is there any constant of an evolving semantic link network? Is there any principle for communication through semantic links? Is there any semantic locality? Is there any limitation of growth?

(4) *Method*. How to discover the implicit semantic links, the emerging semantic communities in the evolving semantic link network, and the principles of evolution?

(5) *Influence*. How does an operation on one set of semantic nodes or links influence the other set of semantic nodes or links? How does this influence propagate? How do different operations or different order of operations influence the network?

(6) *Usage*. What kinds of services the semantic link network can provide? How do users benefit from the semantic link network?

(7) *Knowledge flow*. How does knowledge flow from one node to the other to realize knowledge sharing through various semantic links (Zhuge, 2006)?

The semantic link network on the growing complex data provides an implicit Web for human to act intelligently.

## 3. Multi-Dimensional Space on Data

Dimension is a method for observing, classifying and understanding a space. Among different definitions in various mathematical models, dimension (also called axis) is used to specify points within Cartesian coordinate space with a minimum number of coordinates. Dimension is often used to represent a subset of the attributes of a set of objects or a class of objects in various contexts.

3.1 *Dimension*

Data involve in scale and dimension. Various automatic classification methods and clustering methods are ways to realize one-dimensional classification (McCallum, et al., 1998; Hofman, 1999). If appropriate dimensions can be established on big data, a multi-dimensional space can significantly divide the volume of data (Zhuge, 2008; Zhuge, 2012; Zhuge, 2015).

If the expansion rate of data is $e^n$, a space with at least $n$ dimensions and each dimension having at least $n$ coordinates can be used to manage big data since $n^n \geq e^n$ holds when $n \geq e$. Generally, if the expansion rate of data is $x^n$, $n^n \geq x^n$ holds when $n \geq x$.

If the expansion of data accompanies the classifications of appropriate dimensions, this kind of expansion is controllable. Further, the increment of dimensions does not overload data management. Any point can be accurately located by giving its coordinates at each dimension.

Automatically adapting a multi-dimensional classification space is needed to meet the need of velocity and variety of data. The key operation of managing multi-dimensional classification space is to locate a point in the multi-dimensional classification space that contains big data.

What are the appropriate dimensions? Dimensions represent human understanding of the real world. The dimensions on big data concern the following two aspects:

(1) *The nature of the observed system*. Some dimensions like time are general while other dimensions are specific. The nature of the observed system should be measurable or identifiable such as physical location, weight, and temperature.
(2) *The existing dimensions of the observers' knowledge*. The mapping between the dimensions in users' mind and the dimensions on data enables users to understand data with familiar dimensions and to establish links between familiar dimensions and unfamiliar dimensions. Different users have different dimensions in mind. The formation of a community and its evolution usually accompanies the formation of common dimensions.

3.2 *Multi-dimensional classification space*

Classification is a method of categorizing data. A multi-dimensional classification space can be constructed by classifying a data set from different dimensions. Each dimension takes the form of category hierarchy representing the generation and specialization between categories. A point in the space specifies a category and has a projection at every dimension (Zhuge, 2012).

A dimension may consist of dependent categories, and one dimension may also depend on another dimension in a space without normalization. A space without normal form restriction can only support operations without correctness requirement (like Web browsing). Removing dependent categories and dependent dimensions is a way to normalize a space. Normalization can reduce the volume of data and ensure the correctness of operating the space (e.g., retrieving data and maintaining structure). A set of normal forms has been proposed for normalizing a crude classification space (Zhuge, 2008). A space can focus on a part of objects in a complex system, for example, people in a city.

*It is a task of system analytics to normalize a classification space to guarantee the correctness of operations according application requirements*.

The multi-dimensional classification space has the following advantages in

managing big data:

(1) It supports the exploration of various representations through generalization or specialization from different dimensions.
(2) It is a basic semantic representation that can be easily understood by human with different knowledge structures and can be processed by computer, so it is suitable for managing various representations in different domains.
(3) It can quickly and accurately locate representations by reducing the search space from multiple dimensions.
(4) The category hierarchy of each dimension can localize the change at the lower level while keeping the high-level structure stable when the structure needs to adapt to the representations of the coming new types of objects.

A multi-dimensional classification space can be designed by human according to domain analysis or automatically generated by extracting dimensions from the representations of data. The combination of human design and automatic generation is more effective in complex applications. The computing performance of the multi-dimensional classification space depends on the physical storage and indexing methods.

3.3 *Complex multi-dimensional space and the tasks of analysis*

The multi-dimensional classification space can be extended to a complex multi-dimensional space where dimension can be also a space such as category hierarchy and distance space.

As a complex system, a big country like China generates big economic data everyday, with thousands of dimensions. Selecting independent dimensions can quickly assess the current economic status according to small number of dimensions.

Among all dimensions of a complex system, some dimensions are more important in reflecting the nature of a complex system. Selecting the core dimensions of a complex system like economy is not easy. For example, economic indicators are used to reflect economic status and help analyze economic performance and predictions of future performance. Economic indicators include many indices such as *unemployment rate*, *quits rate*, *housing starts*, *consumer price index*, *consumer leverage ratio*, *industrial production*, *bankruptcies*, *gross domestic product*, *broadband Internet penetration*, *retail sales*, *stock market prices*, and *money supply changes*. Economic indicators can be classified into three categories according to their effect time in economic cycle: *leading indicators*, *lagging indicators*, and *coincident indicators*, each of which consists of several indices, thus forming an indicator hierarchy. These indicators can be regarded as the candidate dimensions of the economy as a complex space.

Some indicators play more fundamental roles. For example, China has used the following three indicators to effectively reflect the economic status of the whole country (named Keqiang Indicator): *electricity*; *railway transportation*; and, *mid-term and long-term loan*. Selecting the fundamental indicators to reflect the economy of a country is an important issue for assessing the economic status of the country. The fundamental indicators should be independent from each other. The fundamental indicators are the candidates of the dimensions of the economic space.

*One task of data analysis is to select a small number of independent dimensions from large number of candidate dimensions*.

*Big data may mislead analysis if data are not hold by the dimensions of the observed complex system*.

A small system can be a complex space. For example, a road can be viewed as a complex space with the following candidate dimensions: *construction*, *structure*, *transportation*, *economy*, and *environment*. The structure may consist of different scales varied from large building blocks to molecular structures. The environment dimension consists of *air*, *water*, *soil*, and *landscape* aspects. Different facets involve in different branches of science and engineering.

As shown in Figure 2, a city is a multi-dimensional complex space where each dimension is also a multi-dimensional complex space. For example, economy dimension is a complex space of multiple candidate dimensions: *resources*, *production*, *consumption*, and *market*. Environment is a complex space with candidate dimensions: *air*, *water*, and *land*. A point in the space represents a category that contains a group of individuals. Every point has a projection at every dimension, which *concerns the measurement on big data*. For example, transportation status concerns real-time big data on vehicles, roads, materials and people. Different spaces have many specific dimensions and principles although they may share some dimensions like time dimension.

*One space can be split into several spaces and several new spaces can join one space*. For example, the space of a society evolves with such operations that some spaces are being split and some new spaces are being joined. A dimension splits when one group of individuals continually interacts with each other much more often than the other group. Two dimensions tend to merge into one dimension when individuals of the two dimensions often interact with each other. The split and join operations may influence dimensions so that data in new spaces may not be accessible from the other dimensions.

*One of the tasks of big data analytics is to find projections from one dimension onto the other dimensions*.

For example, finding the influences of a traffic accident at multiple dimensions (e.g., environment, economic and healthcare) is useful for estimating the total lost. Modern city has been developed into a so complicated system that any discipline is not able to provide necessary knowledge for unveiling its core laws and principles, and that coordinating so many disciplines is so complicated that it is beyond the ability of any individual scientist or any community of discipline. Managing the knowledge of multiple disciplines is becoming a fundamental challenge to science and engineering. This is the basis of managing, analyzing and making use of big data.

*A problem in the development of science is that research of one discipline does*

*not influence other disciplines in time. It leads to the border of knowledge and essential reinvention.* Establishing a multi-dimensional space to enable scientists of different disciplines to share some dimensions so that they can share research in time.

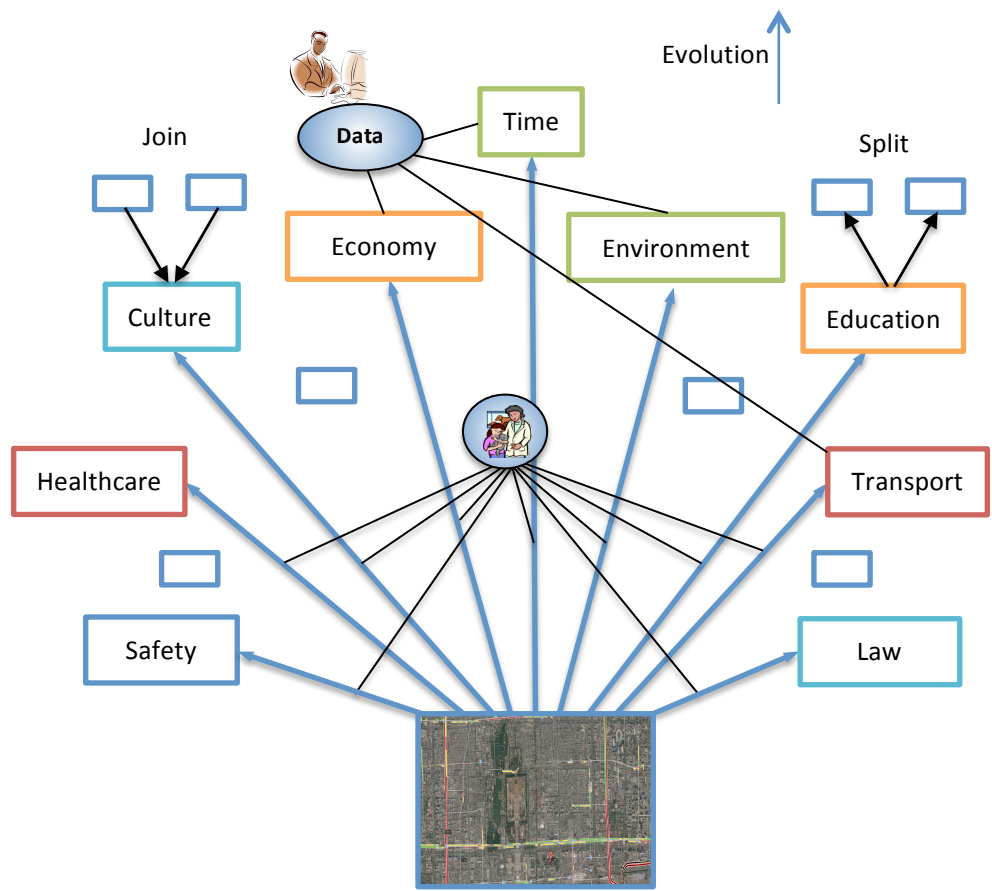

Figure 2. A city evolves a multi-dimensional space with various operations such as joining new dimensions and splitting existing dimensions. Different from other concepts of space and dimension, each dimension specifies data, physical objects and people who interact with each other, with data and with physical objects following specific rules, and consequently generate new data. Different spaces can be generated by selecting different dimensions.

Human are able to establishing dimensions but unable to manage many dimensions. Machines are unable to discover dimensions but are able to help human to find, verify and manage many dimensions.

*Establishing a multi-dimensional space to hold big data generated by the observed complex system is a basic approach to data management and analysis. A key task of data analysis is to select small number of dimensions, to ensure their independency and each of which can specify all data of the space so that all data can be access from different dimensions*.

## 4. Multi-Dimensional Analytics

Current proposals for big data analytics mainly focus on data. Big data is significant only when it reflects the status of the complex system from many dimensions such as time, location, human, value, organization and behavior, which determine the way to organize and operate data.

4.1 *Data operation dimension*

Data operation dimension includes the following operation behaviors:

- *Capturing*, including the analysis of the structure and function of the observed system that generates big data, the design or selection of the instruments or sensors that collect data, the deployment of instruments or sensors, and the integration of data from different instruments or sensors.
- *Structuring*, including pattern discovery, statistic, deployment (data can be deployed onto different connected computing units), indexing, and storing.
- *Managing*, including various management operations on data, which may be distributed.
- *Explaining*, establishing cause-effect link between data and other forms of representations.
- *Exploring*, including top-down refinement and bottom-up generalization from multiple dimensions.
- *Linking*, finding the links between changes at different dimensions.

4.2 *System behavior dimension*

System behavior dimension concerns the trajectory and rules of the observed system, which mainly concerns organization behaviors such as planning, operating, predicting, optimizing, managing, deciding, and coordinating between behaviors.     Behaviors in different spaces follow different rules.

*System behavior analysis focuses on how, from which dimension, and to what extent to enable data to support behaviors to realize the aim and the value of an individual, an organization or a complex system*.

For industrial ecosystems, system behavior analysis concerns the relationships between data and the following aspects (Allenby and Graedel, 2002):

- *Scales*, which concerns enterprise, enterprise community, and industrial ecosystem scales.
- *Aims*, which concern *strategic, tactical and operational aims*. An organization's long-term and mid-term development aims determine its strategy and trajectory, and short-term aims determine its operation.
- *Profit and cost*, which concern such factors as strategies, market, policy, consumers, time, and space. Profit and cost change with the market and development stage. Such fluctuation requests real-time processing of data.
- *Efficiency and impact*, which concern resource efficiency (save resources and money), social impact (creating jobs and donation), and environmental impact (green development model).

- *Opportunity*, which concerns competition (new entrants and products), risks (accident, nature disasters, economic or political crisis), responsibility (social responsibilities for providing safe goods), and environment (controlling pollution) and stakeholders.
- *Sustainable development*, which concerns strategies for improving behaviours to gain higher values or lowering the cost and more positive environmental benefits, e.g., raising bargaining power and enhancing competition power and carrying on green transformation.

The system behavior analysis requests the approach to organizing big data to support analysis of the above aspects.

4.3 *Value dimension*

Value is a dimension of the economy as a space. The value of data depends on the extent that people benefit from using it. A complex system has an image in form of a value network at the economy dimension.

A complex system consists of a network of behaviors such as design, production, marketing, delivering and services in enterprises that generate values. The value of the whole network is greater than the sum of individual behaviors. Data should serve the behaviors that contribute values to the value network (Rayport and Sviokla, 1995). Data should be captured according to the relevancy to the behaviors and stored in the forms and locations that can efficiently support behaviors. The value of data depends on the effect of the behaviors of using data, which may take place at multiple dimensions.

4.4 *Time dimension*

Any system evolves with time. A system shows significant difference in feature, structure and function at different time, at different dimensions or at the same dimension.

Like organisms, social organizations experience a cycle of life: born, grow, mature, decline and ultimately death. The cradle-to-cradle proposal is to extend life cycle by transferring the death stage to a new birth stage through reforming organization and products. Different companies of the same enterprise may have different stages and lengths of life cycle. Different strategies (e.g., shortening the introduction stage, maximizing the growth stage, prolonging the maturity stage, slowing the decline stage, reforming the death stage) and different stages of life cycle request different data supports.

Life-cycle analysis evaluates the status and impact of the observed system (enterprise). If the enterprise is at the introduction stage, data about relevant products (including components), potential competitors, customers, suppliers, market, and constraints like environmental protection should be selected to support product innovation, design and development.

For enterprise at the growth stage, data on new or potential entrants including products (specification, investment, scale, etc.), market (distribution, prices, etc.) and customers (interest, distribution, income, etc.) can be collected from external environment. Different data sets should be selected from the big data according to different strategies (e.g., on product distribution, price, and business network).

4.5 *Human dimension*

*Big data analysis involves in a dual complex system consisting of an observed system (e.g., a factory) and a human system. The two systems interact with each other directly or indirectly, co-evolve and generate data*. The human system consists of two groups: users and people who operate the observed system.

The main purpose of analyzing human behavior is to find, understand and estimate various requirements so as to effectively operate the observed system (e.g., organize resources for production) to provide on-demand products or services. A good service can increase users' values. Creating valuable requirements is an important task of innovative enterprise. Big data about human behaviors provides the sources for discovering, estimating and creating requirements, which drive the behaviors of operating the observed system.

Users in different domains have different habits of using data. Neglecting user behavior analysis often leads to the failure of data management systems. For example, National Cancer Institute's caBIG data integration project wasted ten years' time and millions of funding as reported in (John Boyle, 2013).

User modeling has been a research topic in software engineering and human-machine interaction (Jacobson, et al, 1999; Fisher, 2001). Enabling users to compose their individual interfaces is a solution but this requests rich, standard components, and user-friendly languages for composition and correctness guarantee.

Analysis is a cognitive process. Different analyzers may reach different results on the same set of data. An analyzer may miss some dimensions or factors that influence the effectiveness of using data. Getting feedback from users in time, referring to successful experiences, and discussing with experts during analysis can help make better analysis. To develop a system that can automatically capture the required data with an aim requests the system to equip the observed system with certain knowledge.

In some applications like social network, users constitute a big social system that also needs to be observed and analyzed. As shown in Figure 3, the two systems interact and influence each other to constitute a dual complex system. System analytics should consider the data generated by users, the data generated by the observed system, the correlation between the two data, and the causation of changes.

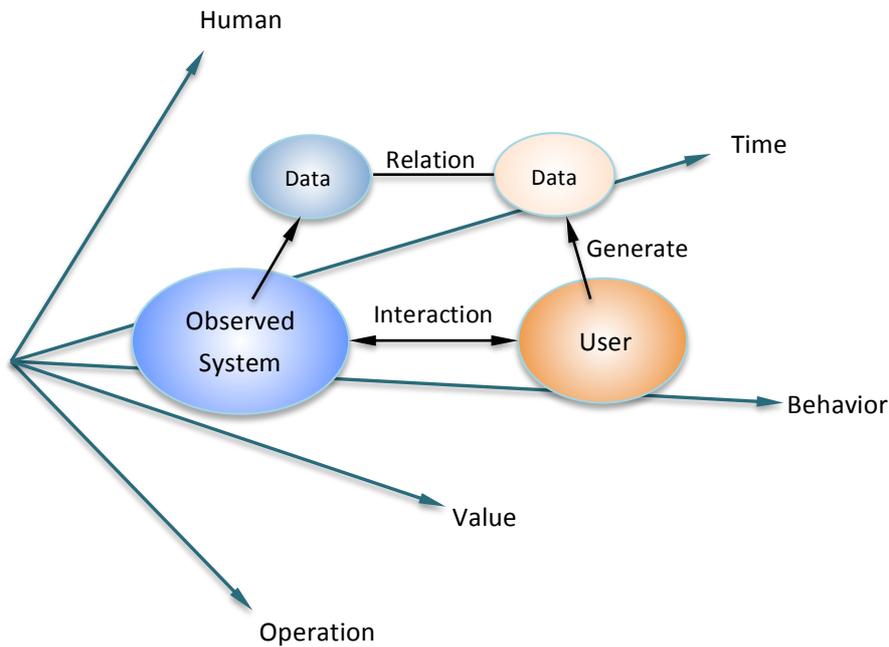

Figure 3. Analysis of the dual system consisting of the observed system and human in a multi-dimensional space.

4.6 *Strategic planning over multiple dimensions — an example*

Analysis over multiple dimensions needs to select suitable models, construct the dataflow and workflow among models, select appropriate data sets for different models, provide a global view on the output, and explain the output.

Figure 4 shows an example of data analysis for enterprise strategic planning, which concerns organization, product, production, marketing, customer relations, and investment strategies. The architecture includes a close-loop data flow through data processing, analysis portfolio management that manages and coordinates various analysis applications such as value chain analysis, life cycle analysis and competition analysis, and strategic planning.  Big data are mapped into a data space by establishing and classifying links.   Different strategies request different data supports.  Different subspaces will be selected from the data space according to different requirements from different analytic applications.

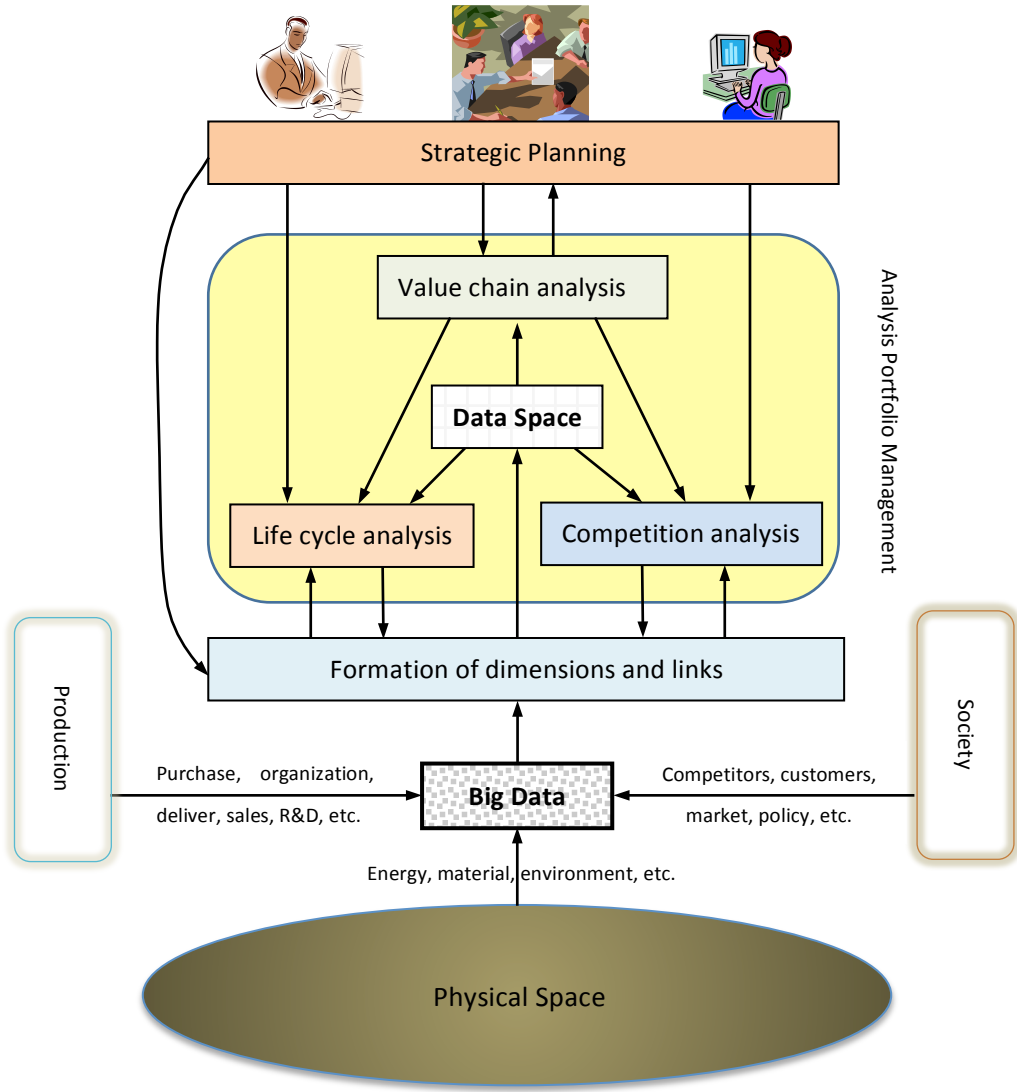

Figure 4. Example: Big data analysis for enterprise strategic planning.

## 5. Unconventional Mapping from Data into Knowledge Space

5.1 *Difficulties*

*Is there a correlational, causal, or reciprocal relationship between knowledge and experience*? This Plato's problem has challenged scientists, psychologists and philosophers for centuries. Nowadays, various representations and interactions in cyberspace have become an important part of experience.

The nature of a class of research in computing area is to transform one form of representation into another form of representation. For example, data mining is to transform the database representation into a different representation (e.g., correlation) (Fayyad, 1996; Han, 2006). Establishing the mapping between different representations is a way to know their expression abilities. The mapping between Resource Space Model, OWL (Web Ontology Language) and database were investigated for integrating different representations (Zhuge, 2008). Representation transformation concerned the isomorphism and homomorphism of information structure and information quantity respectively (Korf, 1980).

It is a natural idea to automatically extract knowledge from data by mapping the representation of data into the representation of knowledge. It is the main research aim of data mining, text mining and data science (Fayyad, 1996). However, *it is difficult for computer to determine and derive the meaning of representations, to calculate meaning, and to explain result*. These difficulties indicate that <u>the correctness of representation is undecidable in computer</u>.

This proposition implies that there is no truth or false for representation. So, it is meaningless to prove the correctness of a representation. This indicates that <u>there does not exist a model that is better than another model in all applications</u>.

For example, many models have been developed for text applications. It is hard to prove that one is better than the other. Any model has its advantages and shortcomings. It is infeasible to find a model that is better than the other in all cases. But, it is feasible to verify that a representation does not satisfy a particular requirement or a representation satisfies a particular requirement, for example, retrieving one node that connects to the other node with a certain relation.

Current representation approaches are limited in ability to reflect the meaning of the complex systems such as interactions in large social networks, countries and cities. One representation may not be transformable into another representation, or the target representation is too complicated to be formally represented. If the meaning of representation and the meaning of mapping cannot be determined, the correctness of mapping cannot be proved. So, it is infeasible to extract knowledge from data by automatically transforming various representations.

*What is between representation and knowledge?*

5.2 *Mapping data space into knowledge space through cognitive system*

Exploring the origin and essence of knowledge and the way to effective knowledge sharing and knowledge management are grand scientific challenges. Knowledge is a core concept of epistemology, cognitive science and artificial intelligence.

Different understandings of knowledge determine different research methods (e.g., symbolism and connectionism) and lead to different results (Newell & Simon, 1976; Newell, 1980; Fodor and Pylyshyn, 1988). Knowledge has been commonly accepted as the most precious property of individual and organization, and research and development on knowledge-based systems attracted a lot of investment, but most systems (including expert systems and knowledge management systems) failed. A key cause is that those systems do not distinguish knowledge from data (Fahey and Prusak, 1998).

Knowledge representation and reasoning is the basis of building various intelligent systems. Five roles of knowledge representation were identified (Davis, 1993): a surrogate for things, ontological commitments, a fragmentary theory of intelligent reasoning, a medium for efficient computation, and a language about the world.

The role of knowledge representation in realizing machine intelligence was questioned from epistemological point of view (Clancey, 1993). The key point is that machines are not able to understand representations like production rules. Indeed, machines process various representations as symbols without knowing their meaning. We cannot expect machines to derive meaningful knowledge from symbols. Brain imaging devices like fMRI can capture data of physical brain but these data cannot reflect motivation and thinking.

Representation involves in structure and semantics. Web semantics is an effort to make machine understandable semantics by creating various Web languages such as RDF (Resource Description Framework) (Berners-Lee, 2001). Semantics is the pathway from representation to knowledge. Interactive semantics emphasizes the importance of interaction in understanding. The following two points are significant in exploring knowledge (Zhuge, 2010; Zhuge, 2011):

(1) Representation indicates semantics that mediates interactions between humans, between machines, and between human and machine.
(2) Knowledge is learned or discovered by cognitive systems interacting with each other in society.

Humans learn or discover knowledge by mapping data into knowledge space through observation, reflection, experience, interactions (including human-to-human, human-to-nature, human-to-machine, human-to-data, and machine-to-machine), thinking (including synthesizing process and various reasoning processes), and representations of different levels. Knowledge in mind evolves through continuous conceptualizing, linking, thinking, experiencing, verifying, rejecting and accepting behaviors, and externalizes through representations (Zhuge, 2012).

Social interactions form and evolve communities for learning, sharing and discovering knowledge. Social networks help build global cooperation that enables people in different regions to share knowledge with each other, accelerate the evolution and differentiation of communities, and record the interaction between people and behaviors.

The internal motion of physical space and social space and the interaction between them generates big data. Data collected through various instruments designed according to different principles are representations of the system to be observed, which can be organized into more understandable forms.

Knowledge develops with forming categories. Some are explicit while some are implicit. Some are commonsenses while some are abstractions. Concepts are basic knowledge components generalized from representation. Computing systems are capable of finding correlation from big data through statistics, which is beyond human expertise. Behaviors such as experiment, computation and analysis on big data help people discover knowledge. However, it is hard for computing systems to automatically generate concepts, principles, laws and methods from data. It is even harder to generate theory from data.

People have been trained in using databases and searching the World Wide Web. *Exploring data is the process of establishing matching between external representation and internal representation*. Experience in the nature and society is the basis for understanding data and generating inspiration from reading data. The fine data of the process of a falling apple may not be able to inspire thinking, while scientists often get inspiration from the real information. This requests an appropriate space to represent data.

*Knowledge is the source and the result of data analysis*. It is hard for data analysis to draw satisfied conclusion from data without domain knowledge. For example, the input-output data between enterprises of a country constitutes resource flow networks and money flow networks. Computer scientists usually can give such measures as rank, centrality, diameter, and community from the sense of graph, but they are hard to draw some domain-specific rules and impacts from the networks. Economists can analyze economic structure, profit, tax, salary, employment, raw materials, and disclose the relationship between different industrial sections and forecast the influences of investment. Ecologists can analyze waste discharge and disposal cost, and disclose the relationship between price and disposal cost. Therefore, *a cognitive system with computing knowledge and domain knowledge is important to discover the evidence or implication in big data to support research, decision and inspire thinking*.

Figure 5 depicts the mapping from the data generated by the observed system in the physical space into the virtual knowledge space through the representation space and a group of cognitive systems that evolve with the interaction with the physical space and the external representation space. The representation space evolves with various operations including management operations like adding new representations to the space and transformation operations such as classification, linking, matching, generalization, specialization, and integration. A common semantic basis such as ontology (Gruber, 1993) and the Interactive Semantic Base (Zhuge, 2010) is necessary for realizing integration and transformation between representations.

The arrows represent the transformation between representations, which can be done by human or physical systems. The interaction between the physical space and the representation space and the interaction between the physical space and the cognitive system are material and energy. Computing systems run in the physical space with the support of material and energy.

The virtual knowledge space is not directly accessible from the external representation space. It is the cognitive systems that discover knowledge through interacting with the physical space, representing information, building concepts, proposing and verifying assumptions in the representation space, and externalizing knowledge with a certain form of representation.

*The cognitive systems essentially self-organize to form some communities of cognition and practice*, where knowledge is shared, verified or rejected. *Reciprocity information (questions, answers, etc.) flows through the members of communities to inspire knowledge discovery*. A distinguished example is the reciprocity flows through Hilbert, Turing and Neumann when they solve the mathematic problem, invented computing model and computing infrastructure respectively. A community of great minds including Hilbert, Turing and Neumann formed fundamental contribution to computing.

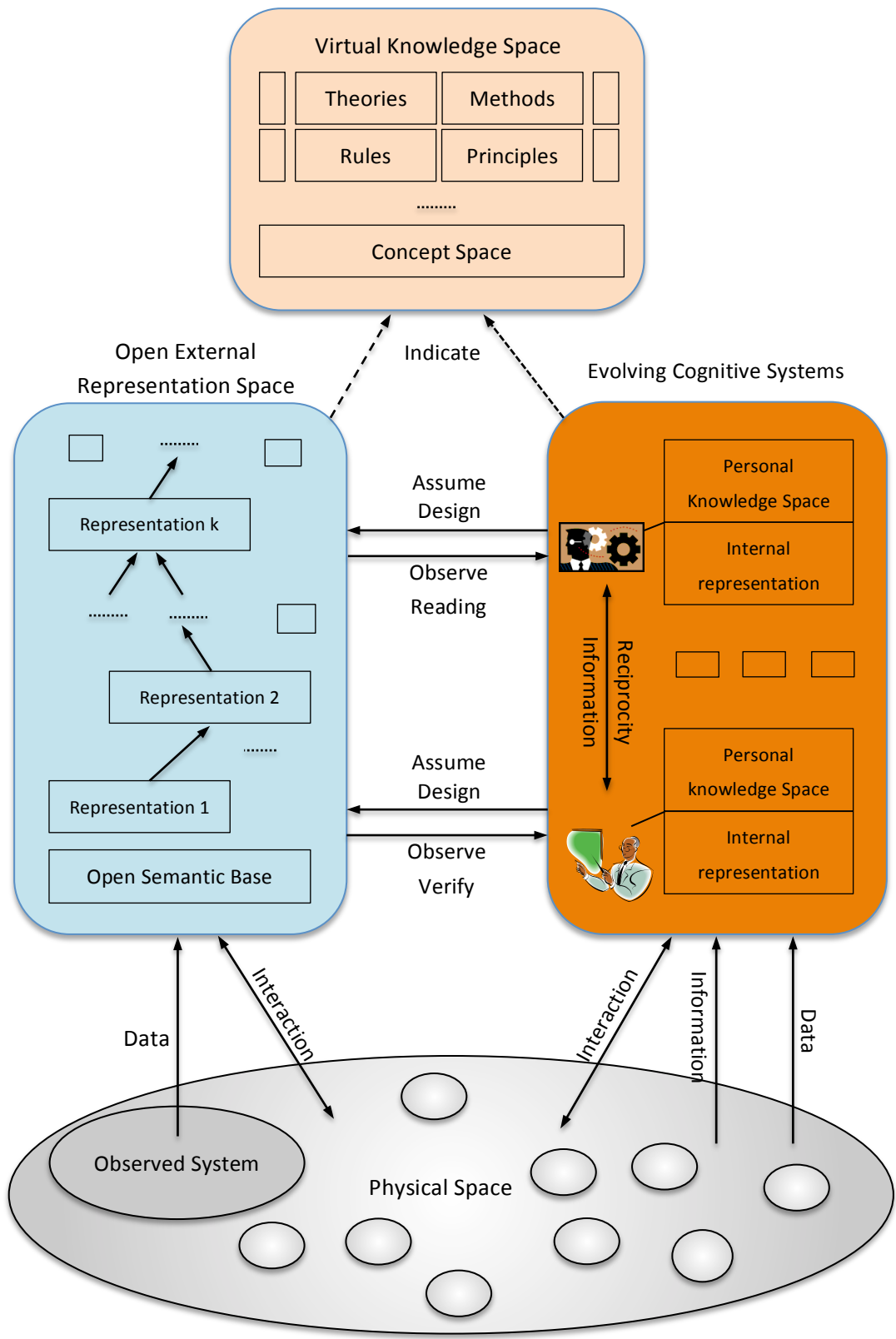

Figure 5. Mapping from external representation space into knowledge space through a group of cognitive systems.

5.3 *From data to concepts*

Data of a certain representation level indicate a certain concept. Words are the basic processing level in most language processing applications such as natural language processing, information retrieval, text summarization, document recommendation, etc.

Concepts are the basic units of representation for thinking. The study of concepts concerns philosophy, psychology, cognitive science, and computer science. Scientists have built various ontologies to facilitate machinery understanding of data.

Human concepts have rich intention and extension, and the formation of concepts is a long-term experience and learning processes. Reading is a process of constructing semantic link network of concepts in mind (Zhuge, 2012). Human understand external representation according to the internal semantic link network of concepts.

*Can we build a semantic link network of concepts in machines*? If we can, machines are able to provide concept-level services like accurate question answering, because the semantic link network of concepts reflects understanding and abstraction.

It is impossible for machines to build concepts that are the same as mental concepts because their natural and social differences. However, it is possible to build cyber-concepts to simulate mental concepts. A way is to define a framework mechanism of cyber-concept and then to enrich it through interaction process. With the semantic link network of words extracted from texts and the semantic link network of cyber-concepts, computers can provide concept-level services by processing texts. As shown in Figure 6, scanning texts according to a certain model can generate a certain information model such as semantic link network of topics that consist of words.

Most previous definitions of concepts are static and passive, and they are mainly for computers to process. In an interactive environment, it is important to define concepts in a multi-dimensional space, which contains some dimensions that facilitate computing (e.g., data structure) and some dimensions that facilitate human understanding (e.g., pictures). The class definition in object-oriented programming can be the basic structure for concepts as it facilitates computers to model class and abstraction, which are the basic features of concepts. The following is a concept space with the following dimensions: structure, services, experiences, rules and sense. A concept in the space has projections on these dimensions.

*Concept-Space:* ( *Structure:* (*attributes*, *classes*, *instances*, *relations*);
 *Services:* (*interfaces*, processes);
 *Experiences:* (*use-cases*, *objects*, *events*);
 *Rules;*
 *Sense:* (*video, picture, language*)).

The structure of concept consists of the attributes of the concept, the classes of the concept, instances and relations. The services provide the interfaces for inputting and outputting information according to some processes. The experiences include use-cases, objects in the physical space, and relevant events. The services can also operate the structure, experiences and rules. The rules are condition-action rules for actions and reasoning. One concept can connect the other concept through various relations, forming semantic link networks of concepts. Potential links could be derived from existing links. With some prior concepts, the semantic link network of concepts could derive new concepts.

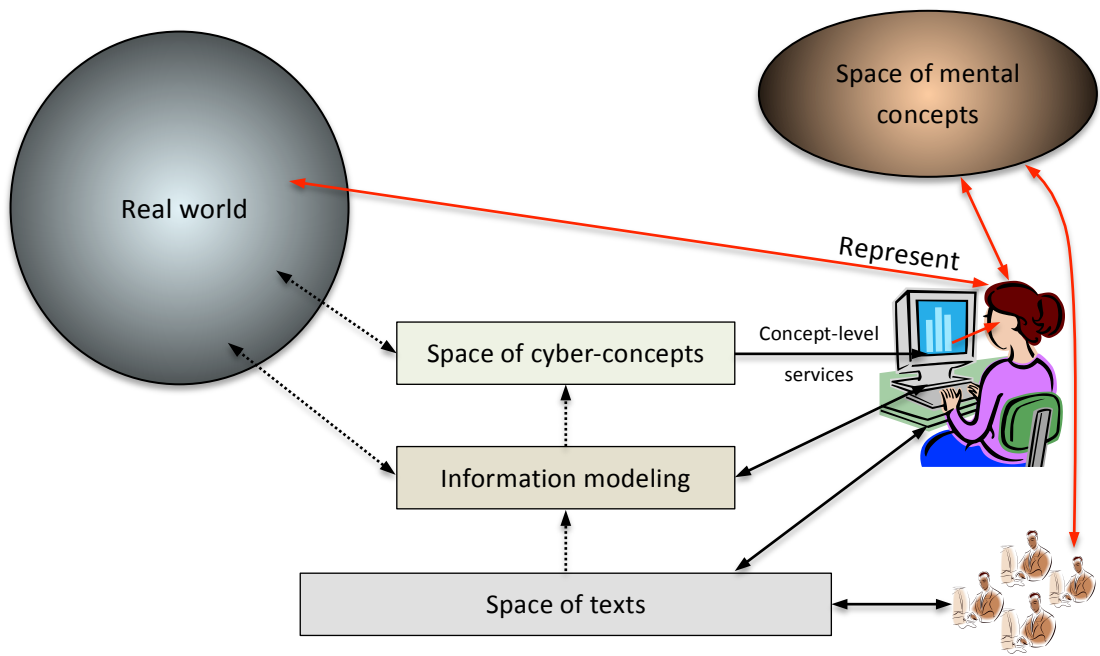

Figure 6. Mapping from the space of texts into the space of cyber-concepts through information modeling.

5.4 *Linking concepts*

Human understanding was explained as a group of mappings between the symbol space, the physical object space, and a multi-dimensional classification space in mind (Zhuge, 2010). As the result of understanding, a semantic image is formed with linking points (concepts) in the multi-dimensional classification space during understanding. Semantic images are emerged and enriched while closing the loops of sensing, controlling, behaving and reasoning through multiple channels (Zhuge, 2012).

Worldview and motivation determine the selection of goals and the selection of representations to be read or observed, and the mapping from the selected representations into concepts. Concepts are enriched and linked while reading. Figure 7 shows that a concept is established by establishing its relations to the real-world object (or event) and the word.

For text understanding, there are different ways of reading. Sequential scanning is one way (Xu and Zhuge, 2013). The following is a way of active reading:

*A reader selects an appropriate text for reading according to motivation, and selects its paragraph and words according to the current goals generated according to the current motivation. The appropriate concept is selected according to the goal and connected to the reading word representation. The goal guides the selection of the next word and the selection of a concept, and links the word to the concept. The relation between two word representations indicates the link between the corresponding concepts. The relation will be also linked to a concept in the concept space according to knowledge (including language knowledge) and experience.*

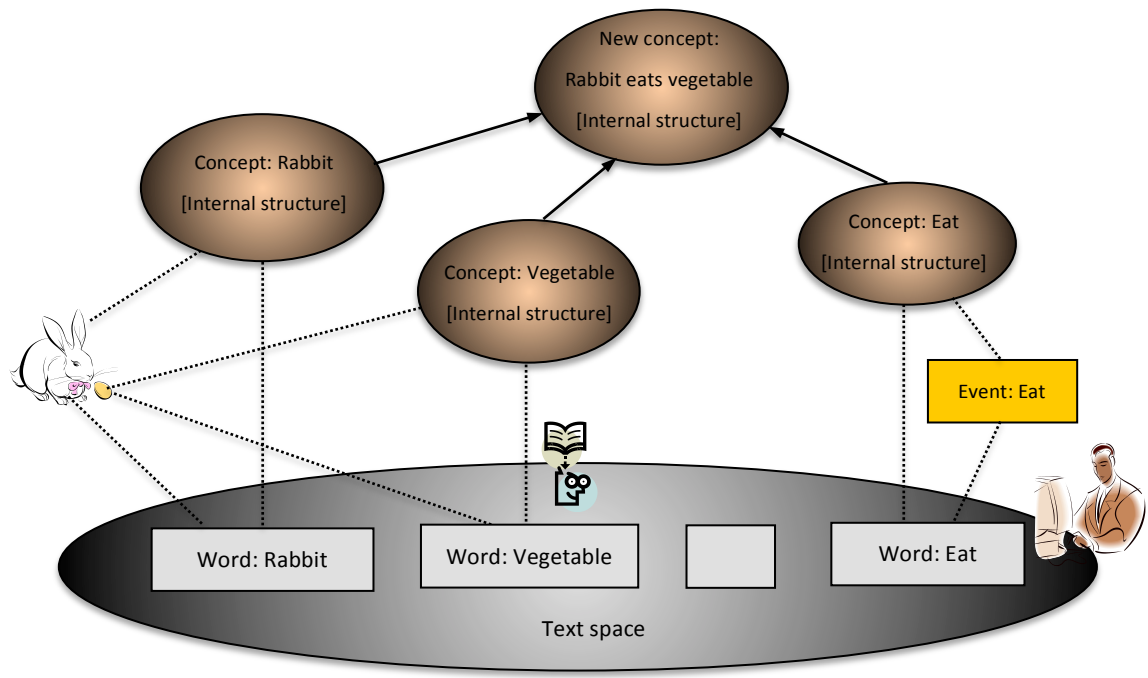

Figure 7. Learning a new concept from existing concepts by connecting concepts and connecting words through reading texts.

5.5 *Mapping from representation into knowledge through complex modeling*

Human can map the representation of the external world into the representation in mind, and map the internal representation into the knowledge space through generalization. Computers are limited in ability to realize this mapping due to the fundamental differences in structure, function and social characteristics.

It is necessary to study an indirect mapping from the external representation into a cyber knowledge space by modeling human cognition. The ability of modeling can be enhanced by adopting the following strategies:

(1) Distinguishing data (including texts), information and knowledge. A major shortcoming of the current data systems, information systems or knowledge systems is that data, information and knowledge are not significantly distinguished.
(2) Separating data processing system, information system, cognitive modeling

system, and knowledge system, and maximizing their capabilities by enabling them to focus on doing one thing, because the capabilities and challenges of data processing systems, information systems and knowledge systems are different.

(3) Coordinating information modeling, cognitive modeling and knowledge space modeling to gain the combined power of building machine intelligence.
(4) Enabling the modeling system to open to human and the Internet to break the limitation of traditional closed systems.

Figure 8 depicts the indirect mapping from representations (e.g., texts) into a cyber knowledge space through information modeling, cognitive system modeling and knowledge space modeling as well as interactions between these systems.

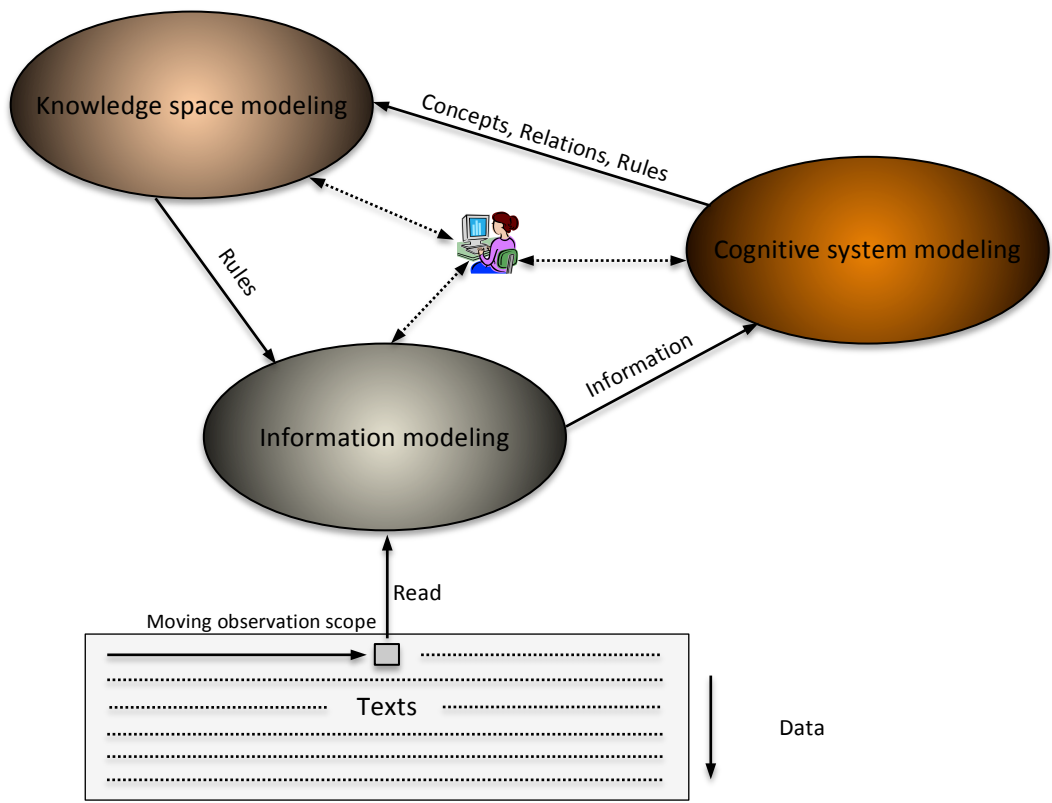

Figure 8. Unconventional mapping from data space into knowledge space through information modeling, cognitive system modeling, knowledge space modeling and interactions — a complex intelligent computing model.

The information modeling system transforms data (e.g., texts) into information with computable information models (e.g., proposition network, cognitive map, probability models and generative models) and measures (e.g., measuring information quantity in text) through reading, or transforms information into text(s) according to information models (e.g., generative model).　 To model the human observation scope in reading process and writing process, an observation scope can be used to dynamically generate information within a certain scale. The small scale is within the scope, the middle scale is within the text, and the large scale contains all texts that have been read so far.  The information modeling system can also input rules from the knowledge space to guide information modeling, and can also coordinate multiple models to make use of the advantage of each model.  The observation scope zooms in and out with scanning to get information of different scales according to rules.

The cognitive modeling system is to discover concepts, relations and rules through various reasoning mechanisms. It receives information from the information modeling system, forms concepts (cyber-concepts) and links them to existing networks of concepts under the guide of the high-level concepts, constraints (e.g., time) and motivation.  It remembers, associates and forgets to evolve the impression of concepts. The cognitive modeling system works with a conscious system and an unconscious system, which interact and compensate each other.  The conscious system consists of memory management, motivation management, and various reasoning mechanisms (analogical reasoning, inductive reasoning, and deductive reasoning), explanation mechanism, and management of various mechanisms. The unconscious system mainly consists of forget, implicit association and meta-cognition mechanisms.  The cognitive modeling system outputs concepts, relations and rules to the knowledge space modeling system for verification and organization. Cognitive map is a kind of representation that enables intelligent behaviors (Tolman, 1948).  Fuzzy Cognitive Map (FCM, Kosko, 1986) is a computing model that can represent cause-effect relations and calculate the influence between concepts. But it relies on human to define the cause-effect relation, and it is limited in ability to compute various semantic relations between concepts.

The knowledge space modeling system consists of several self-contained knowledge systems and mappings between systems. A knowledge system consists of structure, operations, rules and reasoning mechanisms. It verifies the input (concepts, relations or rules) provided by the cognitive modeling system.  If the input can be proved by existing knowledge, the knowledge space accepts the input as knowledge. Otherwise it rejects the input.  Its structure can be modeled by a lattice consisting of high-level general concepts (the top level reflects the worldview) and low-level basic concepts, which forms a particular category of knowledge.  Some concepts are priori while others are posteriori. Human create the priori concepts, which supervise the learning and classification of posteriori concepts. The middle-level concepts are the specialization of the high-level concepts and the generalization of low-level concepts. The structure of the lattice can be enriched through the following steps:

(1) Obtain the category hierarchy by analyzing the open Web contents such as Wikipedia and ODP (Open Directory Project) and existing ontologies, which indicate popular categories.
(2) Inherit a concept and enrich it as the middle-level concepts by the information provided by the information modeling system.
(3) Derive and verify relations between concepts according to rules and various reasoning mechanisms.

The unconventional mapping is open as the information modeling system can also receive new information and models, the cognitive modeling system can receive concepts and rules, and the knowledge space modeling system can receive concepts, rules and structures from external systems including other computing systems, human and society.

*This complex system extends representation (or modeling) and reasoning (or operations) in traditional knowledge discovery approaches, knowledge-based systems and information systems to three modelings: information modeling system, cognition modeling system and knowledge space modeling system as open systems, which is more powerful than separate use of individual models.* The interaction between these systems further strengthens the capability of each system. The capability of the interactive complex system and the capability of the separate systems can be compared by Turing test. The interactive system goes beyond the efforts of time, space and computing ability of traditional computing.

The next problem is how to derive knowledge. An important type of knowledge discovery is problem-oriented (or goal-oriented) as follows:

(1) Input a problem (or goal) into the information modeling system, which selects and scans texts, and then generates information about the problem by using appropriate model.
(2) Input the generated information into the cognitive modeling system to discover new concepts and new relations, and input them into the knowledge space.
(3) The knowledge space accepts or rejects the new concepts and relations through verification. The knowledge structure will be reorganized if necessary. It constructs the local knowledge architecture for solving the problem by using the concepts in the goals to trace the cause-effect chain in the semantic link network of concepts.

5.6 *From correlation to knowledge*

Data mining often finds correlations such as the sales of diapers and beer correlation on Friday nights. These patterns provide experience for making decision or plan (e.g., for sales), but they do not reflect cause-effect relation, which plays an important role in discovering knowledge. Further, data themselves cannot explain correlations.

One class of knowledge is commonsense while the other is great knowledge. From macroscopic, great knowledge can reflect the fundamental principles of the nature or derive a discipline of knowledge that influences the way of thinking or significantly improve human life in the long run. From microscopic, great knowledge can solve a class of problems rather than just an instance of a class of problems. Great knowledge usually has big impact on many dimensions. This implicates that great knowledge should be understandable and adopted by many people. Science essentially pursues the beauty of simple.

Big data reflects the statuses of the observed systems rather than knowledge and thought. It is difficult for machines to automatically discover great knowledge in data because the capability of machines depends on the wisdom of designers. Data analysts can discover knowledge only when they have knowledge about the observed system that generates data. The big data era still needs great thinkers.

*<u>Generating great knowledge needs not only knowledge required to solve the problem but also motivation, believe, insight and imagination, which are all beyond data</u>*.

This is why some people can generate great knowledge while others can only generate experience on the same set of big data. For example, Johannes Kepler's motivation was based on his believe: There must have a geometric reason for holding the planets at the particular distances from the sun, and their orbits should be on spheres. While Tycho Brahe's major motivation was to record the positions of planets more accurately: at least ten times more accurate than the best previous work. Brahe's motivation prevents him from generating great knowledge.

*Different motivations of data analysts determine different sets of dimensions being selected for analyzing data, which generally determines the scope and result of thinking*. This further indicates the necessity of multi-dimensional data exploration.

Kepler's discovery of the law of planetary motion is an example of generating great knowledge from analyzing data. Brahe's observatory data of solar system is commonly regarded as the basis of Kepler's discovery. However, we should not neglect the important roles of his knowledge in discovery: (1) Copernicus' heliocentric theory, which provides motivation and analogical thinking during analysis; and, (2) geometry knowledge, which provides the target of abstraction, meta-model and basic concepts for representing the law. This representation also provides computation basis for further study.

Further, Kepler's law inspired Newton to extend his law of motion to the law of universal gravitation, which in turn verifies Kepler's law. Figure 9 depicts the general formation process of Kepler's discovery through the infrastructure (observatory instruments), representation space, and knowledge space. Managing knowledge for effective data analysis is the key to generating great knowledge from analyzing data.

The following are some implications from the above analysis.

(1) *Great knowledge is not in data*.
(2) *Mapping data into great knowledge needs important data that indicate great knowledge and the ability of mapping data into knowledge through a great mind that can carry out information modeling, cognitive modeling and knowledge management*.
(3) *Knowledge management concerns the coordination between the operations on the external representations, transformation from external representation into internal representation, and the behaviors of discovering, using and sharing knowledge*. The behaviors such as searching and writing are representations in broader sense. Existing knowledge inspires thinking, helps representation, and carries out verification.
(4) *Management of representations can help raise the efficiency of analysis by searching, statistics and visualization*. Kepler spent years to analyze Braher's data, which is too long for many applications such as in business and military. An advanced infrastructure can significantly raise the efficiency and accuracy of management.

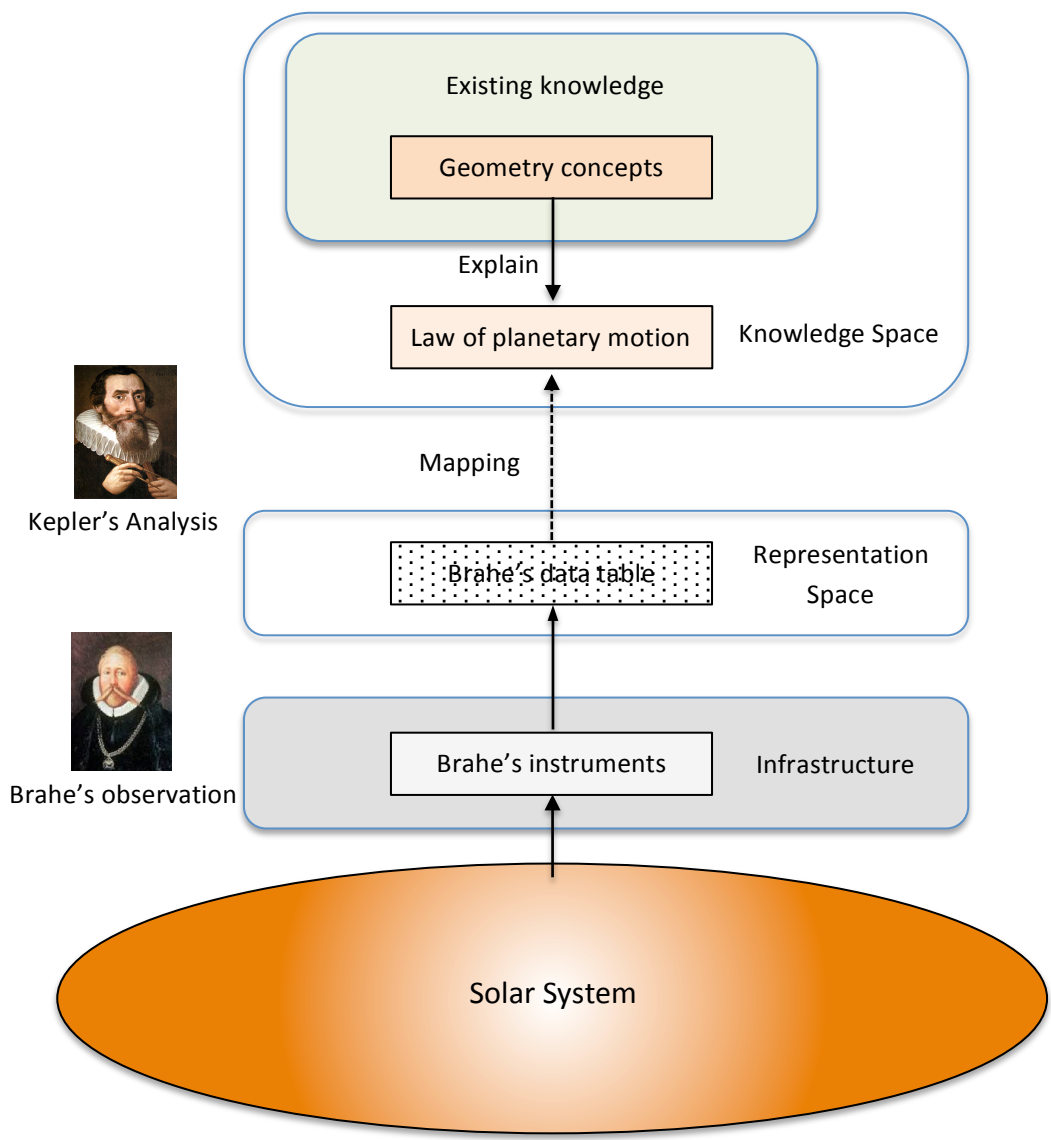

Figure 9. A case study of generating great knowledge through the cooperation of scientists, infrastructure, representation, and knowledge.

5.7 *Human representation and machine representation*

Human have some innate ability to represent some behaviors and languages and to understand representations (McCarthy 2007; 2008). This ability is built through long-term biological evolution and social evolution. This ability can be enhanced through experience and learning. Various representations are formed and evolved through evolution.

Machines can process data with certain representations (including models) designed by human, so the key to processing big data is the representation of data with the "big" features. An interesting idea is to enable machines to automatically generate representations, for example, automatically generating program, data structure, and models of software process (Gries, 1981; Cook and Wolf, 1998; Zhuge, 2012). However, automatic generation systems cannot be automatic. The generative models still request human to design a framework and let computing process to determine its variants. Probabilistic topic models have been widely used for the unsupervised analysis of text, providing both a predictive model of future text and a latent topic representation of the corpus (Blei, 2012). A way to integrate cognitive development, human learning, human abilities, information processing, and aptitude treatment interactions around the transfer of experience and abilities was suggested (Wittrock, 1978).

To understand representations, mental behaviors go far beyond data operation. It is questionable that computational models that perform probabilistic inference over hierarchies of flexibly structured representations can address the deepest questions about the nature and the origins of human thought, e.g., how does abstract knowledge guide learning and reasoning? What is the form of knowledge? And how is knowledge acquired (Tenenbaum, et al., 2011) ?

The nature of human representation and machine representation is different. The representation suitable for machine processing may not be suitable for human understanding, and the representation suitable for human understanding may not be suitable for machine processing.

A solution is to combine human representation and machine representation. The following are possible ways: Incorporating human representation into machine representation, incorporating machine representation into human representation, and separating and coordinating the functions of human and the functions of machine.

5.8 *Knowledge flow through cognitive systems*

Cognitive systems operate knowledge spaces and representation spaces through information modeling. A cognitive system can be modeled as the collections of big number of semi-autonomous, intricately connected agents, with motivation, language, memory, learning, intentions, and metaphors. Different agents can be based on different processes with different aims, ways of representations, and methods for generating results. A society of agents can work together to perform more complex functions than any single agent could (Minsky, 1985, 2006).

A cognitive system has a personal knowledge space and an internal representation space. The knowledge space concerns the environment where the cognitive system operates and rationalizes its behaviors with worldview. The representation space includes the mechanisms that the cognitive system operates, and mechanizes the cognitive system's behaviors with the system view (Newell, 1982). The intelligence and personal knowledge of a cognitive system can be examined by Turing test (Turing, 1950).

Compared to the knowledge flow through the citation links (Zhuge, 2006), Figure 10 depicts a more active knowledge flow from one cognitive system to another with the behavior of linking one representation to another. The external representation space consists of the semantic link networks of various representations.

Knowledge flow is the social process of evolving knowledge while the cognitive system maps one representation into another and carries out internal reasoning with motivation and aim.

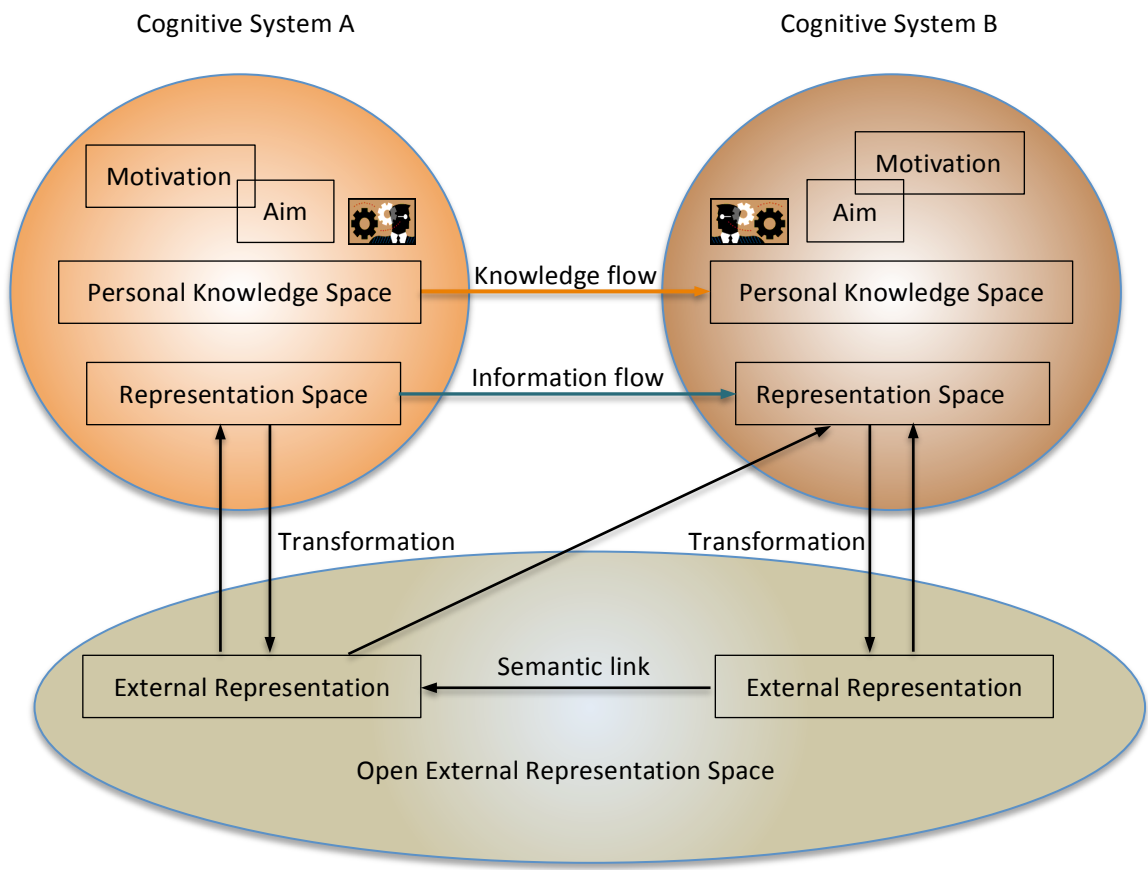

Figure 10. Knowledge and information flow from one cognitive system to another accompany with linking one representation to another.

## 6. Cognitive Cyber-Infrastructure

6.1 *Cyber-infrastructure*

Extending Bush's Memex insight (Bush, 1945), Gray proposed the personal Memex, which can record everything a person sees and hears, and quickly retrieve any item on demand. He also proposed the world Memex, which can answer the questions about the given text and summarize the text as precisely and quickly as a human expert (Gray, 2003).

A big idea in computing history is to transform the thinking computing model — Turing machine into an implementation model by viewing program as data and enabling programs to be fetched and stored in the same way as data (Neumann, 1958). This idea is particular important when storage is expensive. This model has dominated computing architecture since its first description in the First Draft of a Report on the EDVAC in 1945. The Harvard architecture (Mark I developed by IBM in 1944) physically separates storage and signal pathways for programs and data. Efforts to create new architectures can be attributed to the above two types or the combination of the above two types. Dataflow architecture (Dennis and Misunas, 1974) is a significant exploration but it is limited in ability for general-purpose computing.

6.2 *The development of cyber-infrastructure*

Cyber-infrastructure is significantly extended with the development of the Internet, World Wide Web and communication techniques.

The NSF of United States regards the cyber-infrastructure as the converged information technologies including the Internet, hardware and software that support a technical platform where users can use globally distributed digital resources. It provides data acquisition, storage, management, integration, processing and utilization for researchers to conduct scientific explorations.

Grid computing initiated in 1990s was to allow consumers to obtain computing power on demand with analogous in form and utility to the electric power grid (Foster, 2001). A software called Globus was developed to build an adaptive wide area resource environment, and integrated higher-level services that enable applications to adapt to heterogeneous and dynamically changing meta-computing environments.

Cloud computing is realizing McCarthy's vision: "computation may someday be organized as a public utility" (1961). It provides a candidate cyber-infrastructure with efficient storage of data and access to various computing resources and services through high-performance networks with ease, low cost, reliability, and regardless of location and devices (Nelson, 2009). It transformed the paradigm of computing and the business model of computing by making infrastructure, platform, software, and communication as services and shaping a new way to developing and marketing hardware and software. Developers for new Internet services no longer require investment in hardware to deploy their services or human resources for operating hardware and services (M. Armbrust, et al., 2009). IBM developed the computing platform Bluemix as a service (http://www.ibm.com/cloud-computing/bluemix/), which is to support building, running, deploying and managing various applications on the cloud.

The paradigm *programs = data structures + algorithms* has been shifted to distributed *programs = distributed data structures + distributed algorithms*, where data and algorithms can be deployed on massively coordinated machines. For developing reliable, scalable and distributed open-source software, distributed programming environments have been developed to support the processing of big data in a distributed computing environment. The Hadoop software library allows for the distributed processing of big data across the clusters of computers using simple programming models (http://hadoop.apache.og). It is designed to scale up from single server to thousands of machines, each offering local computation and storage. Hadoop has been adopted by Google, Yahoo and IBM for processing big data.

IBM is developing a cognitive computing system to extend both humans and machines, and help humans make better decisions with the help of big data analysis.

Natural language processing, machine learning, image and speech recognition and data visualization are key techniques used to enable people and machines to interact more naturally to extend human expertise and cognition (http://www.research.ibm.com/cognitive-computing). Cognitive applications are planned to develop on the Bluemix. Indeed, it is the time to integrate the research of different areas to make a combined power. So far, it is not clear how human cognition could be developed in this computing environment, and there is no report on combining research on cognitive development and computing development.

The Internet of Things (IoT) extends the Internet to connect various devices such as various sensors and actuators. It is estimated that there will be about 26 billion devices on the Internet of Things by 2020. It has been applied to many applications including environmental monitoring, industrial process management and smart cities. Cyber-Physical Systems try to integrate computing processes and physical process. The Cyber-Physical Society studies the complex space of cyberspace, physical space and social space. The idea of considering both physical domain and virtual domain can trace to Simon's general framework of the science of artificial (Simon, 1969).

Big data captured by various devices deployed in different areas needs a highly decentralized cyber-infrastructure for processing and sharing. The infrastructure extends to the Internet of high-performance computers, servers, personal computers, mobile devices and various sensors. Data from different sources need decentralized processing. Data often need to be processed in real-time by local devices to avoid transmission of big volume of data. Computing processes will be carried out mainly by various devices first and then by local computers, while some can be scheduled to run on high-performance computers. Computing processes can be coordinated to optimize computing resources.

The cyber-infrastructure, along with its resources, users and various connected devices, constitutes a vast artificial environment. Harmonious development of its cyber-infrastructure not only enables people to conveniently access, share and process big data but also ensures sustainable development of society (Zhuge and Shi, 2004). Gödel's incompleteness theorem inspires us to create an open cyber-infrastructure.

6.3 *Big gaps between human and machine*

In addition to obvious structure difference, the following are three key differences between human and machine as depicted in Figure 11.

(1) *Big gap between external representation (e.g., machine representation) and internal representation (i.e., human representation)*. A representation suitable for machine processing is hard to be explained by the representation of human (e.g., language). The reasonability of machine representation (e.g., data structure) is usually demonstrated through particular applications.

(2) *Big gap between machine's symbol processing and human's knowledge processing*. This indicates that it is unreasonable to use human representation (e.g., human classification result) as criteria to evaluate machine representation (e.g., machine classification result), which is widely used in computing areas such as information retrieval and classification. Establishing coordination based on function separation between human and computers is more feasible than pursuing human-level machine intelligence.

(3) *Big gap between the representation of the specialized computing system designers and the representation of the common users.* This determines the gap between the meaning of computing process and the meaning of computing result. This leads to many problems, for example, it is hard for computing systems to explain computing result and to make computing result (e.g., automatic summarization) understandable by common users.

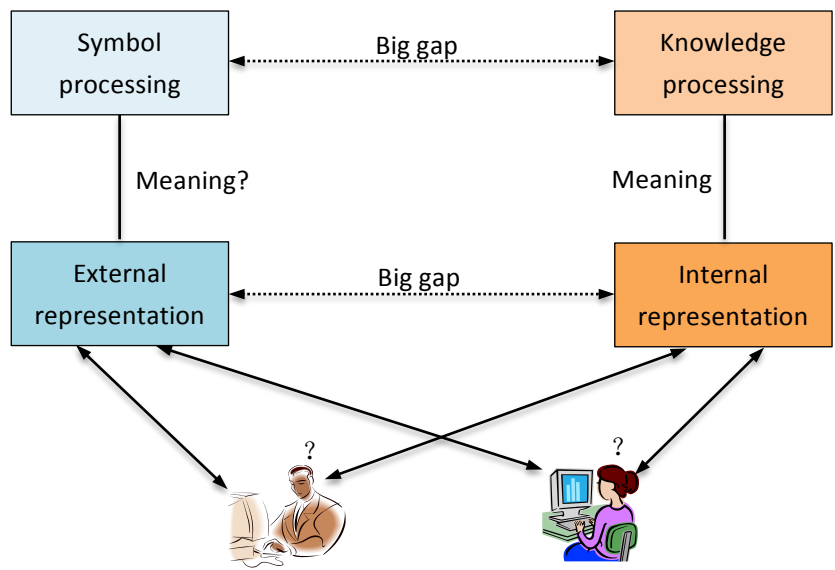

Figure 11. The big gaps between human and machine, which lead to the difficulty of explaining machine representations (e.g., models), understanding representations (e.g., computing results), and evaluating computing process.

Traditional computing researches can be classified into the following streams:

(1) *Improving computing.* It aims at developing better models suitable for machine processing or developing high-performance computers. The underlying reason is that human don't need to understand the representation of computing while just need to use the computing result. Most computer scientists are working along this stream.

(2) *Simulating human to improve computing.* Artificial intelligence is along this stream. This consists of symbolism, connectionism, and multi-disciplinary research. Research often needs to resolve mismatch between human representation and computing representation.

(3) *Simulating the nature to improve computing.* Evolution computing is along this stream. Some researches are on modeling the observed system for better operation or estimation, for example, simulating the generative process of the observed system by probabilistic models and machine learning (Wittrock, 1974; Blei, 2012).

(4) *Simulating human to improve human.* Neumann compared man and computer from the perspective of mathematics and computer (Neumann, 2012), and suggested that: *understand the methods of brain as computation, to recreate these methods, and ultimately to expand its power.* Recreation is often neglected in the research of computing.

(5) *Human-computer symbiosis.* The symbiosis of man and machine can make use of the advantages of both human and computer (Licklider, 1960). For example, computer is specialized in discovering correlations in big data, which is beyond human ability. Human are specialized in thinking and self-motivation, which are beyond the ability of machine.

(6) *Influence on human.* To know how the development of computing influences human is very important for social development. For example, Web search has significantly influenced human thinking (Sparrow, et al, 2011). Many problem-solving issues can be transformed into the issue of searching reasonable solution.

Clarifying different streams can raise the effectiveness of discussion and avoid misunderstanding between scientists working along different streams.

6.4 *Incorporating cognitive architecture into cyber-infrastructure*

Cognitive architectures usually refer to the models of the structure and behaviors of human intelligence (Fodor and Pylyshyn, 1988; Reber 1989; Seger 1994). Traditional cognitive architectures consist of symbolic, connectionist and hybrid. Some architectures are bio-inspired or nature-inspired. A cognitive architecture usually consists of several subsystems reflecting different aspects of intelligence such as action-centered subsystem, non-action-centered subsystem, motivation subsystem, and meta-cognitive subsystem. A proposal of architectures consists of two levels: an explicit knowledge-processing level and an implicit knowledge-processing level, interacting with each other through experiencing and learning.

There are at least four reasons to incorporate cognitive infrastructure into cyber-infrastructure:

(1) *The formation of cyber infrastructure relies on a community of cognition and practice.* Different from traditional information system design or any product design that is to meet requirements, the community should lead practice and have insights on the development of technologies, society and economy. The community grows with the evolution of the cyber-infrastructure. The co-evolution of the community and the cyber-infrastructure constitute a cognitive cyber-infrastructure.

(2) *Simulating the cognitive architecture of the community of practice is a part of creating an intelligent cyber-infrastructure*. The cyber-infrastructure records the evolution of itself and the community of practice. Knowing the cognitive infrastructure of the community of practice helps evolve the infrastructure to bring computing into thinking process. For example, knowing the cognitive architecture of students enables learning systems to dynamically organize appropriate learning materials.

(3) *Emerging cyber-social minds*.  Human minds continuously capture and process information and discover knowledge while cyber-infrastructure stores, manages and processes the data of the observed system and the data of the people who use the cyber-infrastructure.  The cyber-infrastructure can work more effectively on demand if they know human cognitive architecture.

(4) *Incorporating cognitive architecture with behavior-centered subsystem, motivation component and meta-cognitive subsystem into cyber-infrastructure can help understand society and analyze big data*. The Internet of Things enables the cyber-infrastructure to capture data reflecting behaviors and events.

Incorporating cognitive scientists and psychologists who are exploring cognitive infrastructures into the community of cognition and practice for developing cyber-infrastructures can form a new paradigm of cognitive cyber-infrastructure where knowledge can be inspired, studied, discovered, preserved, used and shared efficiently.

From the evolution point of view, an ideal cyber-infrastructure can be created and maintained to provide on-demand services only when the cognitive architecture of the community that studies and develops the cyber-infrastructure reaches a certain cognitive status.  How to manage the cognitive architectures of the community and the cyber-infrastructure is an important research issue.

## 7. Communities of Cognition and Practice

Science, technology and engineering co-evolve with the communities of cognition and practice.  A community of cognition and practice (CCP) usually starts from a small group of people, develops to a large scale, keeps stable development, and inevitably shrinks with the generation of new communities. The life cycle of a community varies with its significance to the development of science and impact (or foreseeable impact) in society.  A community can work with different paradigms to keep sustainable development.  The evolving CCPs constitute the social architecture of the cognitive cyber-infrastructure.

Different from other scientific equipment and research objects, the cognitive cyber-infrastructure enables an idea to attract more interests quickly and to develop quickly by enabling people to communicate with each other more easily. For example, big data has attracted researchers from many scientific fields, government strategic plans and industrial investments, which accelerate the development of the community. Moreover, the cognitive cyber-infrastructure can record not only its own data but also the behaviors of the communities, which can help evaluate and adjust behaviors to ensure sustainable co-evolution.

It is the CCP that inspires and shares knowledge through communicating with each other and through sharing, using, operating and analyzing data. Knowledge can be on data, on the evolving architecture, on the physical space and on themselves.

Inspiring and managing knowledge of the CCP with the support of big data in the cyber-infrastructure is a new paradigm of scientific research.

The cognitive cyber-infrastructure co-evolves with the CCP including scientists who study the cognitive infrastructure and discover scientific knowledge, engineers who maintain the cyber-infrastructure and contribute technologies, users who contribute commonsense and data, policy makers who use it and contribute policy, and funding institutions that provide funding for research and development. This actually extends the man-computer symbiosis (Licklider, 1960) to a community-infrastructure symbiosis.

Figure 12 depicts the cognitive cyber-infrastructure consisting of the physical space, cyber-infrastructure and society. The physical space provides material, energy and information for the operation of the cyber-infrastructure and the society.

Society consists of individual, organization, various CCPs and social factors such as motivation, behavior and value. A CCP evolves with the scale of people, social network and cognitive architecture with people joining and leaving. People involve in diverse roles such as scientist, user and engineer interact with each other, and one role can play other roles. One person can participate in different communities. Both individual user and organization can play a role. The members of the IT CCP communicate with the members of the other communities to get requirements and share knowledge. This is critical to the adoption of a cyber-infrastructure in these communities. Many data management systems in some areas like life science and health failed due to the lack of this communication and knowledge sharing. Communities co-evolve with the cyber-infrastructure through various behaviors such as planning, design, development, maintenance and use. During evolution, new CCPs may be generated from merging existing CCPs.

The information architecture of the cyber-infrastructure consists of the following six layers:

(1) *The decentralized data capture and transmission architecture*. The Internet-of-Things is a candidate architecture for connecting various machines and devices to capture and transmit data.
(2) *The decentralized data storage architecture*, which can efficiently manage distributed data structure.
(3) *The decentralized complex processes consisting of computing, control and communication*.
(4) *The open, decentralized interactive semantic base*, which makes the sense of various representations.
(5) *The multi-dimensional information space*, which represents information at different abstraction levels and different scales. The multi-dimensional classification space, semantic link network and linked data can be the candidates.
(6) *The decentralized interactive processes*, which support various open social networks.

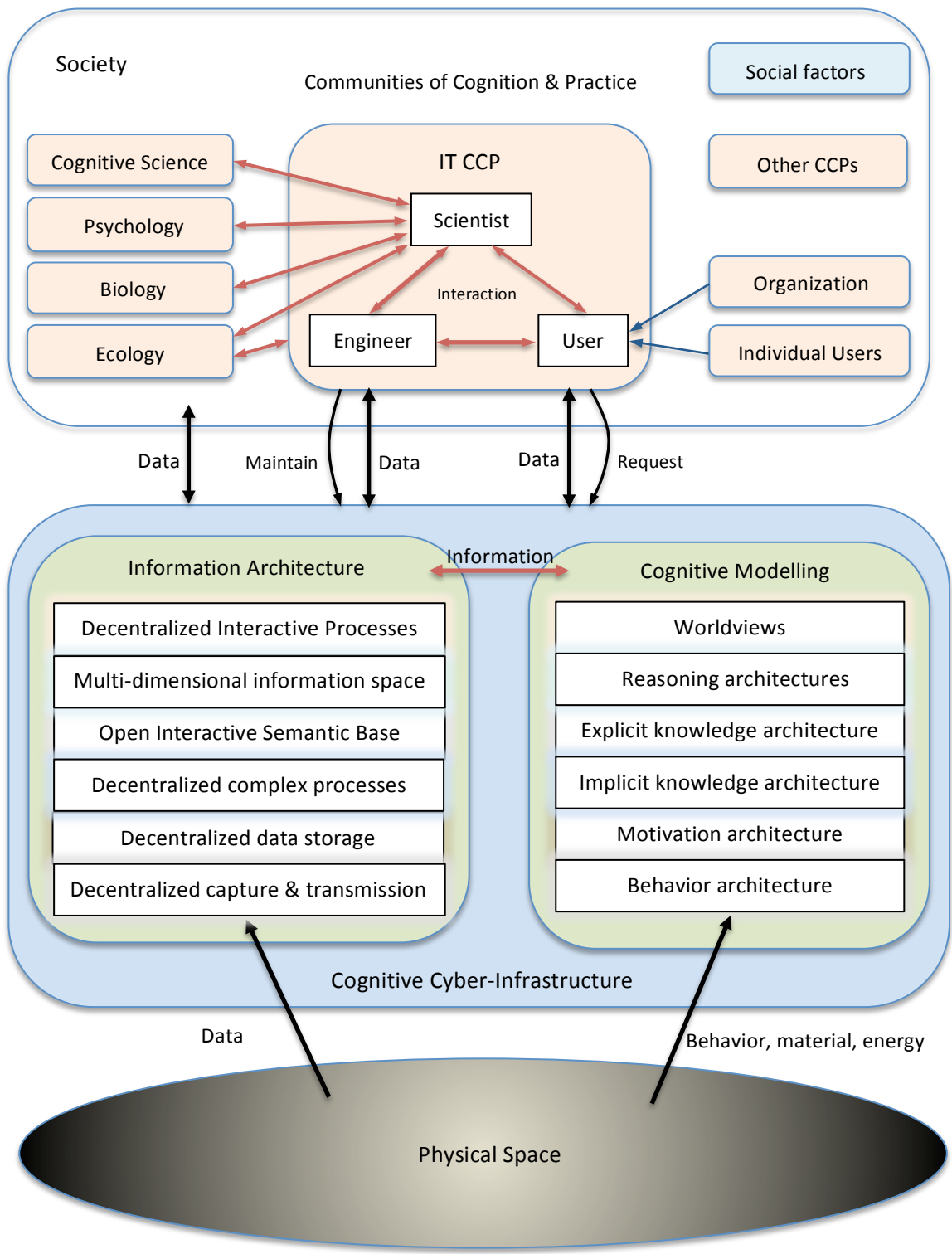

Figure 12. Cognitive cyber-infrastructure.

The cognitive modeling mechanism consists of the following six layers:

(1) The behavior layer, which includes various application behaviors such as decision, planning, optimization, production, sale, etc.
(2) The motivation layer, which generates and manages motivations.
(3) The implicit knowledge layer, which manages the semantic structure that implicates knowledge.
(4) The explicit knowledge layer, which manages the knowledge that can be directly used and shared.
(5) The reasoning architecture, which is responsible for deriving and justifying knowledge according to the lower cognitive layers.
(6) The worldview layer, which manages the top categories that determine the concept hierarchy of knowledge.

The implementation needs a new method of representation, which should go beyond symbol representation.

Figure 13 depicts the structure of the general CCP consisting of three layers: (1) the underlying social interaction network; (2) people who interact through the social networks; and, (3) an implicit social cognitive architecture. The cognitive architecture is formed and evolved with interactions between members through social networks. The implicit architecture facilitates the sharing of knowledge between members through organizational mechanisms including various forums (including physical and online conferences, workshops, and publications), work groups, and protocols with agreed strategies. The domain knowledge consists of concepts, problems, principles, rules and methods, which are understandable throughout the community. The IT knowledge consists of the strategies, the aim and the trend of information technology, the standards of various data structures, interfaces and services, and the scope and the features of information techniques, which are understandable throughout the community.

Society consists of individual, organization, various CCPs and social factors such as motivation, behavior and value. A CCP evolves with the scale of people, the evolution of the social network and cognitive architecture with people joining, interacting and leaving. People involve in diverse roles such as scientist, user and engineer who interact with each other, and one role can play the role of the other. One person can participate in different communities. Both individual user and organization can play a role. The members of the IT CCP communicate with the members of the other communities to obtain requirements and share knowledge. It is critical to adopt a cyber infrastructure in these communities. Many data management systems in some areas like life science and health failed due to the lack of this communication and knowledge sharing. Communities co-evolve with the cyber-infrastructure through various behaviors such as planning, design, development, maintenance and use. During evolution, new CCPs may be generated from merging existing CCPs.

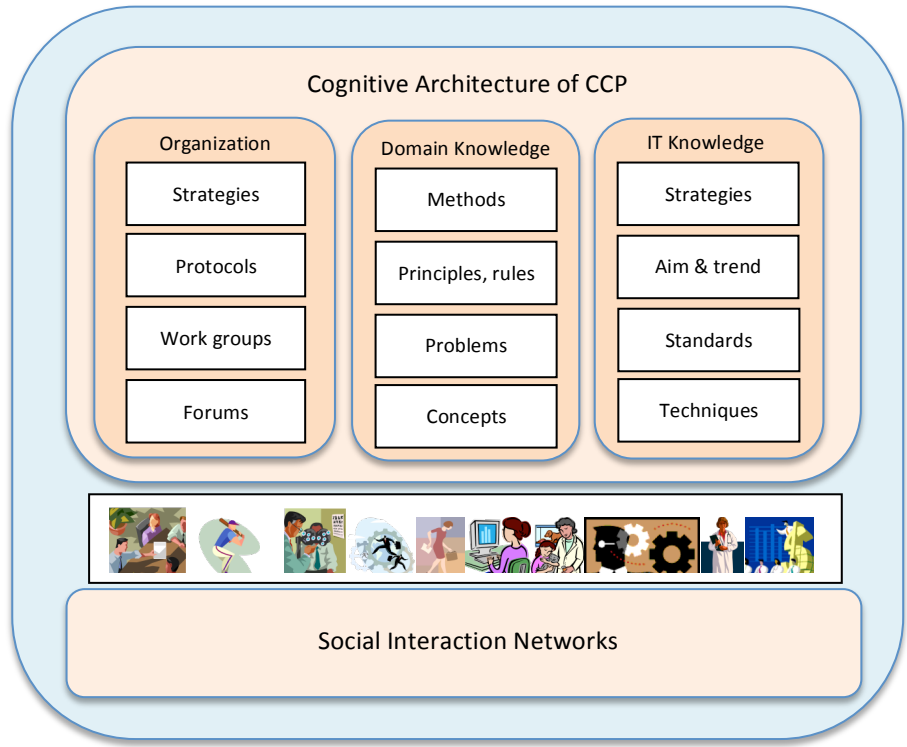

Figure 13. The structure of the Community of Cognition and Practice.

## 8. New Paradigm of Science

*8.1 The evolving paradigm of science*

Science studies the nature as objective reality with some fundamental assumptions about the nature of the universe, e.g., objective reality exists without depending on mind, and the nature has regularity and natural cause (Kuhn, 1970). Establishing coherence between scientific activities and discoveries is a way to build a self-contained scientific system.

A scientific paradigm refers to the scientific achievements that provide model problems and solutions for a community of practitioners during a particular period (Kuhn, 1970). It concerns the observed objects, the formation of questions, research method (including experiment method and equipment), and the interpretation of the results.

Paradigms tend to shift in mature fields. At the end of the 19th century, Einstein published special relativity, which challenged the rules of Newton mechanics that had dominated physics for over two hundred years. The new paradigm reduces the old paradigm to a special case (Newtonian mechanics is a model for the slow speed physical space).

Science paradigm experienced a spiral development from empirical research (describing natural phenomena) to theoretical research through modeling and generalization. Debate between empiricism and rationalism accompanies the development process. A statistician George Box's claim "essentially, all models are wrong, but some are useful" represents the empiricism (Box, 1989). This coincides Karl Popper's argument that the central property of science is falsifiability — every genuinely scientific claim is capable of being proven false, at least in principle (Popper, 2004). Indeed, models are the simplifications of the real world and restricted by the inventors' recognition of the real world. But the important thing is that models are the representation of the thoughts of inventors. Models will be verified, accepted or rejected through practice. Some models are very useful in many areas, some are useful in one area, some are useful in some special cases, and some are wrong.

As the $2^{nd}$ paradigm of scientific research, modeling has limitation, because model is the simplification of the real world, and models are limited within the knowledge of individuals who create them. The changing real world often makes models unsuitable for real applications.

8.2 *Science on data, concept, motivation, thinking, knowledge and interaction*

Data is the basis of experimental science. Data generated from various devices are extending human sensation and complementing direct scientific observation. Big data provides more facts for scientists to analyze and test scientific assumption. However, data need to be represented in an appropriate form to be processed by computers.

It is human who motivate scientific exploration. Thinking plays the key role in developing sciences, especially establishing scientific concepts and theories. Requirements from social development and science development force the generation of motivation. The curiosity of scientists is an important driven force of forming scientific problems, which cannot be generated from data. The generation of scientific theories needs not only experiments but also rational thinking. How to provide real-time services for scientific exploration and thinking according to computing on big data is a research issue.

Scientific research activities have been greatly influenced by the Internet. With the cognitive cyber-infrastructure, scientific research will carry out in an environment of *document-to-document interaction* (e.g., citation, co-author, etc.), *document-to-human interaction* (e.g., writing, reviewing, publishing and commenting), and *human-to-human interaction* (e.g., cooperation). Various types of interactions generate new problems, inspire ideas, and share knowledge. The evolving interaction environment has become a new driven force of science.

The rapid development of computing technology has greatly helped scientists in simulating complex phenomena, thus research has gradually relied on data. Accidently, computer science has experienced the development from theory (Turing model) to technology (building common-purpose computer), from technology to science (formal methods), and to pragmatism (applications) and empiricism (statistics has become an important modeling method).

However, the essence of scientific research is knowledge-intensive rather than data intensive. Scientific research heavily relies on the formation of motivation, concepts, insights and thinking, which rely on mind.

If big data cannot completely reflect the nature, it may generate misleading result, even if just one factor is missed. In many cases, complete data is neither easy to be collected nor guaranteed. Sensors cannot detect some phenomena such as symptom of patients and psychological activities and thinking. On the other hand, people may not know what is missing when collecting data. So, data should be collected with clear goal and plan of use, otherwise data make no sense no matter whether it is big or not.

An e-Science platform connecting various sensors to the Internet can greatly help collect, reserve and share data through research cycles. Correlations exist in the data space. Data will be observed, processed, and used in scientific papers by scientists. Scientific papers will be read by other scientists, cited by the other papers, and commented by other scientists. Scientists participate in various events, cooperate with each other, and form and evolve social networks. Comments will be correlated via papers, readers and writers. Scientists form new knowledge through reading and synthesizing papers and comments. Scientific research will become more efficient and inspiring than ever in the environment of diverse interactions as shown in Figure 14.

Moreover, scientists can observe the evolution of the e-science platform and their own research behaviors. There is no doubt that e-science with big data will accelerate scientific research. But this does not mean we can neglect other paradigms. Previous paradigms will still play an important role in scientific research. Sometimes, great ideas are generated in great minds inspired by just a simple phenomenon. As the fourth paradigm, big data exploration enriches the means for scientific research.

Various interactions establish explicit or implicit links in the interactive environment, which enables cognitive systems to obtain information and generate knowledge according to up-to-date information. Citing (including commenting) is a kind of interaction that could attract potential citations because (1) new citations could incur explicit citation links between papers, implicit comment links between papers and researchers, explicit cooperation links, and implicit knowledge flows between researchers, which form small worlds of research; (2) scientists (including authors and reviewers) could be inspired from reading and writing new contents (papers and comments); and, (3) the quality of papers could be improved with the evolution of authors' knowledge through interaction. This explains some phenomena, for example, rejected papers often gain citations once published after improvement.

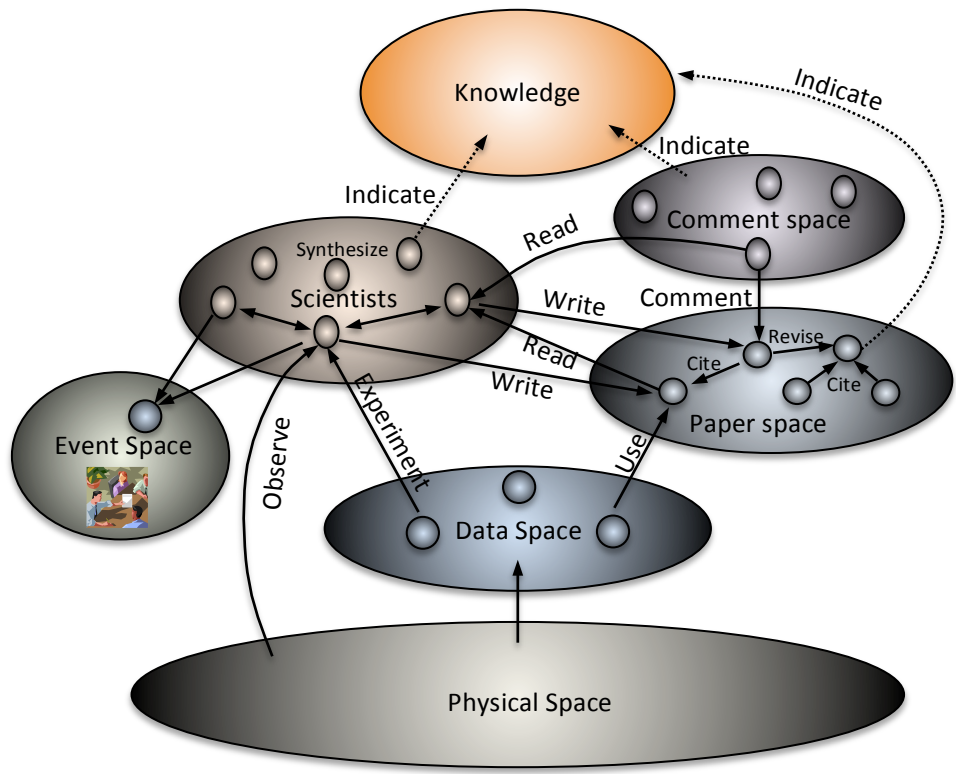

Figure 14. The interactive paradigm of scientific research.

## 9. The Nature of Big Data Computing

9.1 *The computing nature*

Turing solved the decision problem of Hilbert by creating a Turing machine and proving the impossibility of designing a Turing machine program that can determine infallibly within a finite time whether or not a given Turing machine will eventually halt (Turing, 1937). Turing machine simulates the basic operations of human problem-solving processes. It reflects the nature of computing because it is the abstraction of all problem-solving processes.

However, if problem is unclear, unaware, or unable to be represented, problem-solving is not meaningful. Humans live in the world with many unformulated, unclear, unaware and unconscious things, which cannot be computed by Turing machine. *All problems on big data, once represented for computing with Turing machine, belong to traditional computing problems*.

*The nature of big data computing is to go beyond Turing machine to form a new paradigm of computing*. *The basic problem of big data is to formulate problems from big data*.

As shown in Figure 15, human live with the computable world and the un-computable world. The computable world contains things that can be represented. The un-computable world contains many unrepresented things (things that have not been represented or cannot be represented in languages), unknown things (things that exist but unknown), unclear things (things that cannot be clearly represented), and unconscious things.

Big data provide a new condition to enable some un-computable things computable. Big data can be regarded as a kind of representation. In traditional computing, human formulate a problem and obtain the solution from a computing process.

*Big data computing can work when human do not know the problem and recommend problem and solution to human*.

The big data computing process consists of the following three steps:

(1) Collect data from the observed system and input the data into the computing process and then output a problem.
(2) Input the problem and then output the solution.
(3) Recommend <*problem*, *solution*> to the user who does not know the problem before.

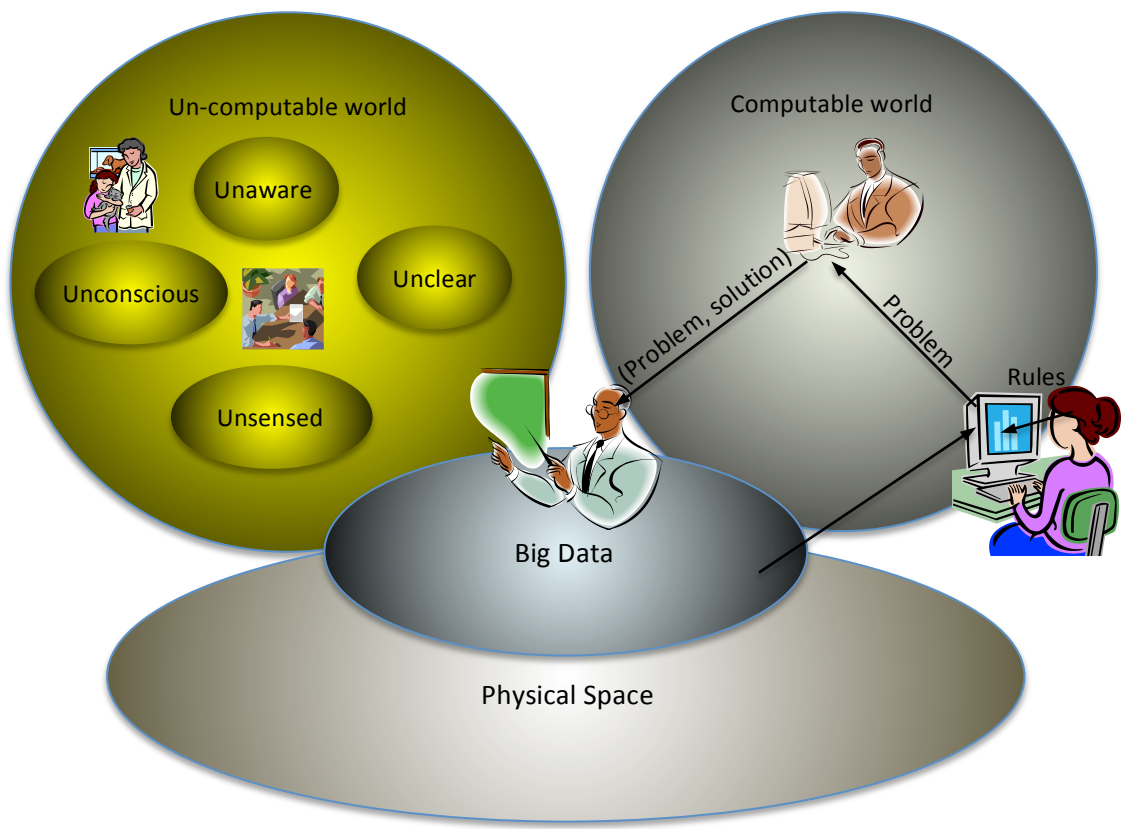

Figure 15. Big data computing with the computable world and the un-computable world.

The following is an example of diagnosis: (1) collect big data of human behaviors through various sensors, detect abnormal patterns, determine potential illness, interview the potential patients, and then determine illness; (2) find the treatment of the illness; and, (3) recommend <*illness*, *treatment*> to the patients who are unaware potential illness.

*The big data computing transfers traditional computing paradigm "find solution" into the paradigm "find problem"*. The above step (1) needs rules assigned by human to guide problem discovery, sometimes interaction between human and machine is needed.

It is important to know the capability of big data. The following propositions unveil the limitation:

(1) *Big data do not increase completeness*. It is difficult (even impossible) to get complete data of a complex system.
(2) *Big data do not lead to accuracy* (in reflecting the nature of the observed system or in understanding a concept). The accuracy depends on the data of the fundamental features of the observed system. Analysis may be misled if only one dimension of the complex system is taken as the global system.
(3) *Big data is not a necessary condition for solving hard problems that rely on algorithms*. Big data provide new conditions for the conventional approaches to process data.
(4) *Big data do not reflect rational thinking*. The way to represent and process data is different from the way to represent and process knowledge in mind.
(5) *Big data neither reflect richer semantics nor help understanding*. For example, a mature scientific field contains growing big volumes of papers, with regard to limited study time of people. Big data technologies are unable to help students learn the knowledge of a scientific field quickly. Scientists have to do research through reading small number of papers. Automatically recommending appropriate papers needs in-depth study of knowledge, understanding and language, which big data cannot help much. To obtain a good (human-level) summarization of growing documents is a hard problem (no great progress with half century exploration in this area). Big data cannot help solve this problem either.
(6) *Big data cannot prove conjectures*. For example, we can enumerate big number of facts (4=2+2, 6=3+3, 8=3+5, ……) to verify the Goldbach Conjecture but we still cannot prove it.

9.2 *Problem-driven, data-driven, and data-based problem driven*

Identifying scientific problems is more important than solving problems. Theoretical research is mostly driven by problems. Current scientific paradigms rely on scientists to raise problems. Only insightful scientists can raise important problems.

Big data provide more facts for scientific exploration. Before the invention of the Internet, scientists can only access small number of references (i.e., small data), so repeated research often happened. Nowadays, scientists can search big number of references, which can significantly reduce the possibility of repeated research. However, with continual expansion of research papers, scientists are getting harder and harder to review the big number of related papers.  How to find related work and summarize them is an important research in big data era, but this kind of big data problem can be solved by traditional computing paradigm.

*The big data of the observed system and the big data of previous research actually indicate scientific problems that need to be explored.  Discovering problems from the big data of various spaces will be a new computing paradigm in this big data era*.

The following are some approaches to raise problems:

(1) Identify relationships (e.g., detecting co-occurrence phenomena identifies symbiosis relationship), verify relationships, and unveil the impact of the relationships in the system.
(2) Make analogy between the existing problem of one system and the problem of another system, for example, identifying industrial symbiosis by applying bio-symbiosis to industry, and developing man-computer symbiosis by applying bio-symbiosis to the computing system.
(3) Generalize existing problems to propose a more abstract problem with the help of the category hierarchy.
(4) Specialize a problem for simplification.
(5) Discover differences (e.g., in definitions, approaches, solutions, etc.) to the same problem.
(6) Detect the limitation of existing solutions by using data.
(7) Synthesize the existing problems.
(8) Complete knowledge by answering "Why?" and "How?".

Bayesian network is a means of reasoning on causally related hypotheses (problems) (Heckerman, et al., 1995).

9.3 *Beyond Turing test*

Turing test has some shortcomings, for example: it assumes that human is wise and computer is fool, therefore computer is not able to lie even if it is more powerful in some aspects (e.g., recite); it regards human as the standard of intelligence while human is not able in some aspects; and, it regards human and computer as close systems, both human and computer are isolated and cannot interact with other people and machines, and external resources (data) are not available.

The natural differences (e.g., in structure, process and function) between human and machine determines that using human behaviors as the standard of evaluating machine is inappropriate.  Turing test has been often misused in evaluating the performance of computing systems.  For example, researchers often use human classification of texts to evaluate machine classification of texts.  The following analogy helps judgment: If we use monkey to replace computer, we cannot say monkey's classification is good if it matches human's classification because monkey's view may be different from human's view due to their natural difference.  Similarly, if we use monkey to replace human to do classification, we cannot say computer's classification is good if it matches monkey's classification.  For statistical computing on big data, we cannot say that computer is more intelligent than human if human cannot find any rule in big data.  *It is more appropriate to compare machines with machines and compare human with human*.

*Finding valuable problems needs insight*. So, finding problem is usually more important than finding solution in science to a certain extent. The ability of finding valuable problems is an important kind of intelligence.

The effectiveness of mapping big data into knowledge can be tested by the following criteria:

(1) **The ability of obtaining knowledge**.  If a system cannot answer a question but it can do within the given time after inputting the data (including tables, texts, images, videos), then we say that the system has obtained certain knowledge from the data and it has the ability of processing the data.  The more questions it can answer, the more knowledge it has obtained from the data.

(2) **The ability of using knowledge**. If a system can propose a problem within the given time after inputting the given data, then we say that the system has the ability of using knowledge to find problem in data.  The more problems it can propose, the stronger ability it has.  The value of the problem depends on its importance and influence in the semantic link (e.g., cause-effect) network of knowledge.

(3) **The ability of proposing problem and solving problem**.  If the system can propose problem on the given data and solve the problem within the given time, then the system has the ability of obtaining knowledge and using knowledge.

9.3 *Analogical mapping*

Analogical reasoning is a basic mechanism of learning concepts from experience. It is a representational mapping from a known source into a target (Hall, 1989). Analogies are usually inspired by semantic cues. The induction of a general schema from analogs facilitates analogical transfer (Gick and K.J.Holyoak, 1983).  The semantic link network of concepts provides rich semantics for analogical reasoning (Zhuge, 2012).

Figure 16 depicts two types of analogical reasoning for mapping bio-symbiosis (source) into two targets: the industrial symbiosis domain (upper part) and the man-computer symbiosis domain (lower part).  If there exists an isomorphism between the source domain and the target domain such that the relations in the source domain can be mapped into the relations in the target domain, the source solution could be mapped into the target domain with high possibility.

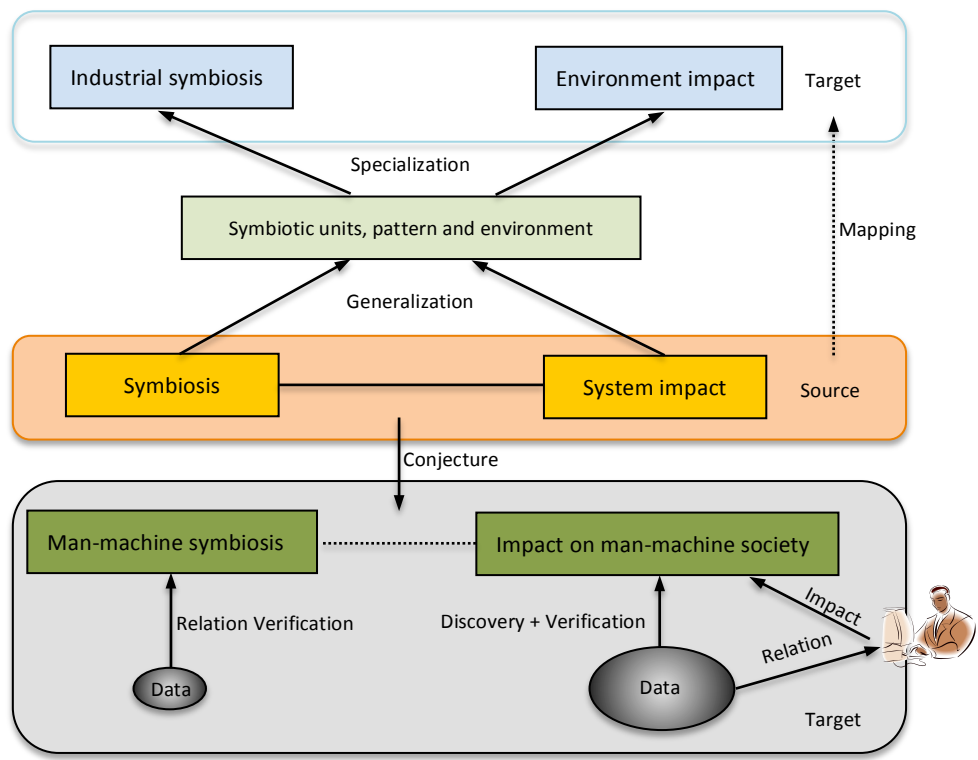

Figure 16. Mapping from the existing domain into new domains.

If there is no such an isomorphism, analogy will try to generalize the source as a general symbiosis (including symbiotic units, pattern and environment), and then specialize the source into the target.

The lower part depicts the type of analogy with conjecture and verification. This type is useful when there is no isomorphism between the source and the target, and the source is hard to be generalized. It consists of the following steps:

(1) Propose the conjecture of the problem (e.g., man-machine symbiosis) in the target domain according to the source (bio-symbiosis relation) and the concepts of the target domain.
(2) Verify the conjecture by examining the existing relations and concepts in the domain.
(3) Propose the conjecture of the solution according to the source solution and the concepts of the target domain.
(4) Discover the relations in the target domain, and examine the relations in the conjecture of the solution.
(5) Examine the impacts of these relations on the other relations in the target domain according to the data generated from the domain. Big data in the target domain help examine the impact.

Big data can help analogy with discovering relations, proposing problems, and verifying conjectures.

*Big data computing will be a new paradigm of exploring unknown world*. Big data computing will accelerate the development of information technologies and the synthesis of existing technologies to develop new technologies, which will transform current paradigms.

9.4 *From big data computing to open interactive computing*

Data is no longer isolated in the global networking environment. Accessing data involves in many interaction processes. In many cases, problems and programs are not predefined. One of the characteristics of interactive computing is that the computing process is constructed during interaction (Zhuge, 2010).

An open interactive computing model consists of interactions of different levels:

(1) *Data-to-Data interaction*, through structuring (various data structures and semantic links) and transforming. Some implicit links may be derived from the interaction.
(2) *Concept-to-Concept interaction*. Concepts self-organize to emerge and evolve semantic images according to a pre-defined goal and the rules for self-organization.
(3) *System-to-system interaction, e.g., interaction among information modeling system, cognitive modeling system, and knowledge modeling system according to the rules for coordinating these systems*.
(4) *Human-to-data interaction*, e.g., access, read and write. Human knowledge will be updated through interaction.
(5) *Human-to-human interaction*, e.g., cooperative study and research, for human-level knowledge sharing.

## 10. Summary

The big data surge emerges with the symphony of the ongoing revolution of technologies, the 4$^{th}$ paradigm of science, the 4$^{th}$ industry revolution, the globalization of economy, and the transformation of cities, which leads to profound social reforms. Big data research tries to use and develop information technologies to extend human sensory to form a more realistic view of the world.

From the macroscopic point of view, big data surge reflects the current development stage of science, technology and engineering that any discipline is not able to provide necessary knowledge for recognizing the rules of complex systems or solving complex problems. Acquiring new knowledge requests the fusion of sciences, technologies, engineering and methods.

From the system point of view, big data reflect the status of a complex system. The fundamental problem is that humans have limited knowledge to establish accurate model for some complex systems such as society and economy because individual knowledge is too narrow to fully understand the complex systems. Big data provide more facts for understanding those systems.

From the data point of view, big data research concerns the following three issues:

(1) *management*, which concerns collection, verification, organization, storage, and operations; (2) *exploration*, which is to discover various relations and patterns in data and understand system behaviors on data; and, (3) *services*, which concern the development of services on big data including question-answering, recommendation, summarization, etc.

An evolving cognitive cyber-infrastructure can provide the state-of-the-art technologies for enabling data-intensive research and inspiring knowledge generation and innovation with big data distributed in different institutions. The Community of Cognition and Practice and the Cognitive Cyber-Infrastructure interact and co-evolve to provide a better environment for minds to discover knowledge.

The development of computer emerges various research methodologies: rationalism like Turing machine, empiricism like data mining, social constructionism like World Wide Web and Wikipedia, and pragmatism like utility computing and cloud computing. The transformation from big data to knowledge needs a multi-dimensional methodology that absorbs and fuses the advantages of rationalism, empiricism, social constructionism, and pragmatism (Zhuge, 2012).

Big data surge arouses the requirement for new scientific and engineering methodology. The driving force is the desire of the knowledge that can transform society. The study of unconventional mapping from big data into knowledge space invokes rethinking of traditional philosophical problems (e.g., the Plato's problem), scientific problems and technical problems towards a new methodology.

**Acknowledgement**. The author thanks the Chinese Academy of Sciences and Aston University for the cooperative research environment and time provided for this work. The author thanks Dr. Xiaoqing Shi in Chinese Academy of Sciences for inspiring discussion on section 1.6, section 4.2, section 4.3, section 4.4, section 9.2, and section 9.3. The author was invited to present ACM Distinguished Lectures on this topic at several international conferences, Chinese Academy of Sciences, British Computer Society, and a number of universities in China, the UK, and the USA.

**Hai Zhuge** is a Distinguished Scientist of the ACM and a Fellow of British Computer Society.

Homepage: http://www.knowledgegrid.net/~h.zhuge.

Email: zhuge@ict.ac.cn